\documentclass[authoryear,3p,review,12pt]{elsarticle}

\usepackage{amssymb}
\usepackage{amsmath}
\usepackage{algorithm}
\usepackage{algorithmic}
\usepackage{hyperref}
\usepackage{graphicx}
\usepackage{multirow}
\usepackage{subfigure}
\journal{Computers \& Operations Research}

\begin{document}
\begin{frontmatter}
\title{A Tabu Search Algorithm for the Multi-period Inspector Scheduling Problem}

\author[hust]{Hu Qin}
\ead{tigerqin@hust.edu.cn, tigerqin1980@gmail.com}

\author[sysu2]{Zizhen Zhang\corref{corl}}
\ead{zhangzizhen@gmail.com}

\author[sysu]{Yubin Xie}
\ead{xyb.0606@gmail.com}

\author[cityu]{Andrew Lim}
\ead{lim.andrew@cityu.edu.hk}

\address[hust]{
School of Management, Huazhong University of Science and Technology,\\
No. 1037, Luoyu Road, Wuhan, China
}

\address[sysu2]{
School of Mobile Information Engineering,
Sun Yat-Sen University, \\ Tang Jia Bay, Zhuhai, Guangdong, China
}

\address[sysu]{
School of Information Science and Technology,\\
Sun Yat-Sen University, Guangzhou, China \\
}

\address[cityu]{
Department of Management Sciences,
City University of Hong Kong,\\
Tat Chee Ave., Kowloon Tong, Kowloon, Hong Kong
}

\cortext[corl]{Corresponding author at: School of Mobile Information Engineering,
Sun Yat-Sen University, Tang Jia Bay, Zhuhai, Guangdong, China. Tel.: +86 13826411848.
}

\begin{abstract}
This paper introduces a multi-period inspector scheduling problem (MPISP), which is a new variant of the multi-trip vehicle routing problem with time windows (VRPTW). In the MPISP, each inspector is scheduled to perform a route in a given multi-period planning horizon. At the end of each period, each inspector is not required to return to the depot but has to stay at one of the vertices for recuperation. If the remaining time of the current period is insufficient for an inspector to travel from his/her current vertex $A$ to a certain vertex $B$, he/she can choose either waiting at vertex $A$ until the start of the next period or traveling to a vertex $C$ that is closer to vertex $B$. Therefore, the shortest transit time between any vertex pair is affected by the length of the period and the departure time. We first describe an approach of computing the shortest transit time between any pair of vertices with an arbitrary departure time. To solve the MPISP, we then propose several local search operators adapted from classical operators for the VRPTW and integrate them into a tabu search framework. In addition, we present a constrained knapsack model that is able to produce an upper bound for the problem. Finally, we evaluate the effectiveness of our algorithm with extensive experiments based on a set of test instances. Our computational results indicate that our approach generates high-quality solutions.
\end{abstract}


\end{frontmatter}

\section{Introduction}
\label{sec:i}
This paper studies a new manpower routing and scheduling problem faced by a company that procures products from over one thousand suppliers across Asia. The company places orders with a large number of suppliers and must inspect the goods at the factories of the suppliers before shipment. Therefore, the suppliers are required to make inspection requests with the company when the ordered goods are ready for delivery. An inspection request is characterized by the workload, the inspection site and the time window within which the inspection can be started. In turn, the company dispatches a team of professional quality inspectors to perform all on-site inspections. In order to facilitate coordination between inspectors and suppliers, the inspections could only be carried out during working hours (e.g., 8:00 am to 6:00 pm). Usually, a weekly schedule is created to assign inspectors to requests for the upcoming week. The company has a stable of in-house inspectors, each having a specified weekly workload limit, and the unfulfilled inspection requests are outsourced to external agencies with additional costs. After receiving their weekly inspection schedules, the inspectors depart from the regional office and will not report back until they have performed all their assigned inspections for the week. More specifically, they leave the regional office on Monday, visit a set of inspection sites and return to the regional office on Friday or some earlier workday. In each workday, an inspector generally travels to diffident locations, completes several inspections and finds overnight accommodation (i.e., hotel) in the vicinity of his/her last/next inspection site at or before the end of the office hours. The objective of the problem is to assign as many inspection workloads as possible to the stable of in-house inspectors while satisfying all the above-mentioned practical constraints.

We call this problem the {\em multi-period inspector scheduling problem} (MPISP), which can be viewed as a variant of the multi-trip vehicle routing problem with time windows (VPRTW) \citep{Azi2010Exact,Macedo2011Solving}. There are four main features that distinguish the MPISP from the multi-trip VRPTW. First, the scheduling subjects, e.g., vehicles or inspectors, are not required to return to the regional office every workday. Second, at the end of each workday, each scheduling subject must stay at one of the vertices for recuperation. Third, each vertex can be visited more than once. If the remaining time of the current period is insufficient for an inspector to travel from his/her current vertex $A$ to a certain vertex $B$, he/she can choose either waiting at vertex $A$ until the start of the next period or traveling to a vertex $C$ that is closer to vertex $B$. The vertex $C$ is called a {\em waypoint}, which only acts as the intermediate point in a route. Fourth, the objective is to maximize the total inspected workload rather than to minimize the number of inspectors used and/or the total distance traveled.

In this study, we propose a tabu search algorithm to solve the MPISP. This algorithm employs a tailored fitness function consisting of three lexicographically ordered components, a local improvement procedure with tabu moves, an ejection pool improvement process and a perturbation phase. The contributions of this study are fourfold. First, we introduce a new and practical multi-period manpower routing and scheduling problem that considers multiple working periods. Second, we provide an effective tabu search algorithm that uses a set of problem-specific neighborhood search operators. Third, we construct a constrained knapsack model that can produce an upper bound for the MPISP. Fourth, the comprehensive experimental results on a large number of test instances show the effectiveness of our approach.

The remainder of this paper is organized as follows. We first provide an overview of related research in Section \ref{sec:lit}. In Section \ref{sec:pd}, we then give a formal definition of the MPISP. In Section \ref{sec:spp}, we describe an approach of computing the shortest transit time for any pair of vertices with any departure time. Our proposed tabu search algorithm is detailed in Section \ref{sec:aits} and the constrained knapsack model is presented in Section \ref{sec:ub}. Section \ref{sec:ce} reports the experiments results and Section \ref{sec:con} concludes this study with some closing remarks.

\section{Related work}
\label{sec:lit}
The MPISP is one type of manpower scheduling problems. Scheduling staff members is a traditional research area; example problems include the nurse rostering problem \citep{Cheang2003Nurse}, the technician planning problem \citep{ernst2004staff} and the airline crew rostering problem \citep{Kohl2004Airline}. As for the manpower scheduling problems that involve creating routes for staff members, we refer the reader to \citet{Li2005Manpower,Tang2007Scheduling,Zapfel2008Multi,Cai2013Tree,zhang2013memetic}.

Since each inspector has to perform inspections at different locations, the MPISP is essentially a variant of the vehicle routing problem \citep{toth2002vehicle}. One of the defining characteristics is its objective of maximizing the total inspected workload. Two previously studied problems with similar objective are the team orienteering problem with time windows (TOPTW) \citep{vansteenwegen2009iterated,vansteenwegen2011orienteering,Labadie2012Team,Hu2014Iterative} and the vehicle routing problem with time windows and a limited number of vehicles ($m$-VRPTW) \citep{Lau2003,Lim2007}. The $m$-VRPTW is an extension of the TOPTW with the consideration of vehicle capacity and customer demands. These two problems both aim to determine a set of routes that maximizes the total reward of the vertices visited during a single period with a distance or duration limit. The multi-period planning horizon of the MPISP is related to the periodic vehicle routing problem (PVRP) \citep{gaudioso1992heuristic,Hemmelmayr2009Variable} and the multiple trip vehicle routing problem (MTVRP) \citep{Battarra2009Adaptive}. However, the MPISP is quite different from the PVRP and MTVRP. In the PVRP, each customer requires a certain number of visits within the planning horizon, and two types of decisions are involved in the planning, namely determining the visit days for each customer and the routing plan for each time period. The PVRP and MTVRP both require that each vehicle must return to the depot at the end of each period.

Another defining characteristic of the MPISP is the consideration of multiple working periods. Working hour regulations have recently received increasing attention from some researchers studying vehicle routing problems. \citet{Savelsbergh1998Drive} proposed a dynamic and general pickup and delivery problem in which lunch and night breaks must be taken into account. \citet{xu2003solving} applied column generation based solution approaches to solve a pickup and delivery vehicle routing problem that involves a set of practical complications, such as heterogeneous vehicles, last-in-first-out loading and unloading operations, pickup and delivery time windows, and working hour restrictions by the United States Department of Transportation. Similarly, \citet{goel2009vehicle,goel2010truck} and \citet{kok2010dynamic} investigated combined vehicle routing and driver scheduling problems under the European Union regulations for drivers.

The MPISP problem can be viewed as a natural extension of the orienteering problem with hotel selection (OPHS) \citep{Divsalar2013Variable,Divsalar2014Memetic}. In the OPHS, a scheduling subject can visit a set of vertices each with a score and find accommodation at a given set of hotels. The tour is divided into multiple trips, each with a limited duration and starting from and ending at one of the hotels. The objective of the OPHS is to determine a tour that maximizes the total collected score. The OPHS is a variant of the traveling salesperson problem with hotel selection (TSPHS) \citep{Vansteenwegen2012Travelling,Marco2013Memetic}, which aims to serve all vertices with the minimum number of connected trips and the minimum total travel distance. The common characteristic of the above three problems is the involvement of hotel selection.

\citet{Tang2007Scheduling} and \citet{Zapfel2008Multi} introduced two manpower routing and scheduling problems that involve maximization of profits, multiple periods and working hour restrictions. However, their problems require that the trip in each period must start from and end at the depot. Most recently, \citet{zhang2013memetic} proposed an inspector scheduling problem which is very similar to our problem. Their problem differs from our problem in the following four assumptions: (1) each vertex can only be visited at most once; (2) if a vertex is visited by an inspector, its inspection request must be fulfilled by this inspector; (3) an inspector reaches a vertex and completes the corresponding inspection task in the same period; and (4) each vehicle stays at the last served vertex at the end of each period and begins the trip of the next period from that vertex. By ignoring these four assumptions, the MPISP is more difficult but practical.

\section{Problem description}
\label{sec:pd}
The MPISP is defined on a directed graph $G= (V, E)$, where $V =  \{0, 1, \ldots, n\}$ is the vertex set and $E = \{(i, j): i, j \in V, i\neq j\}$ is the edge set. Vertex 0 represents the depot location and $V_C=\{1, \ldots, n\}$ denotes the locations of $n$ suppliers. Each supplier $i$ is characterized by a location $i \in V_C$, a workload $d_i$, a required service time $s_i$ and a time window $[e_i, l_i]$. For notational convenience, we assign $d_0 =0$ and $s_0 =0$ for the depot. Each edge $(i, j) \in E$ requires a non-negative traveling time $t_{i, j}$, where the matrix $[t_{i, j}]$ satisfies the triangle inequality.

We are given a set $K$ of $m$ homogeneous inspectors, each of which has a workload limit $Q$ and can only work within a set $P=\{1, \ldots, w\}$ of $w$ {\em working periods} (or called {\em working time windows}). For any period $p \in P$, $a_p$ and $b_p$ $(a_p < b_p)$ are its starting and closing working times, respectively, and $b_p - a_p$ equals a positive constant $T$ that is not less than $s_i$ for any $i \in V$. An inspector can arrive at vertex $i \in V_C$ prior to $e_i$ and wait at no cost until the service of supplier $i$ becomes possible. All inspectors must leave the depot after $e_0$ $(e_0 = a_1 = 0)$ and return to the depot before $l_0$ $(l_0 = b_w)$, where $[e_0, l_0]$ is called {\em depot time window}. At the end of each period, each inspector is not required to return to the depot but has to stop traveling and stay at one of vertices. Moreover, service cannot be interrupted, i.e., if the service of some supplier cannot be completed before the end of a period, it must be restarted in the later periods. Each vertex can be visited more than once while each supplier can be served by at most one inspector, so some supplier locations can be used as {\em waypoints}. The objective of the MPISP is to construct $m$ inspector routes to complete as many workloads as possible while respecting depot time window, workload limit, supplier time windows and inspector working time windows. We provide a mixed integer programming model for the MPISP in Appendix A.

\begin{figure}[!h]
\begin{center}
\resizebox{12cm}{!}{\includegraphics{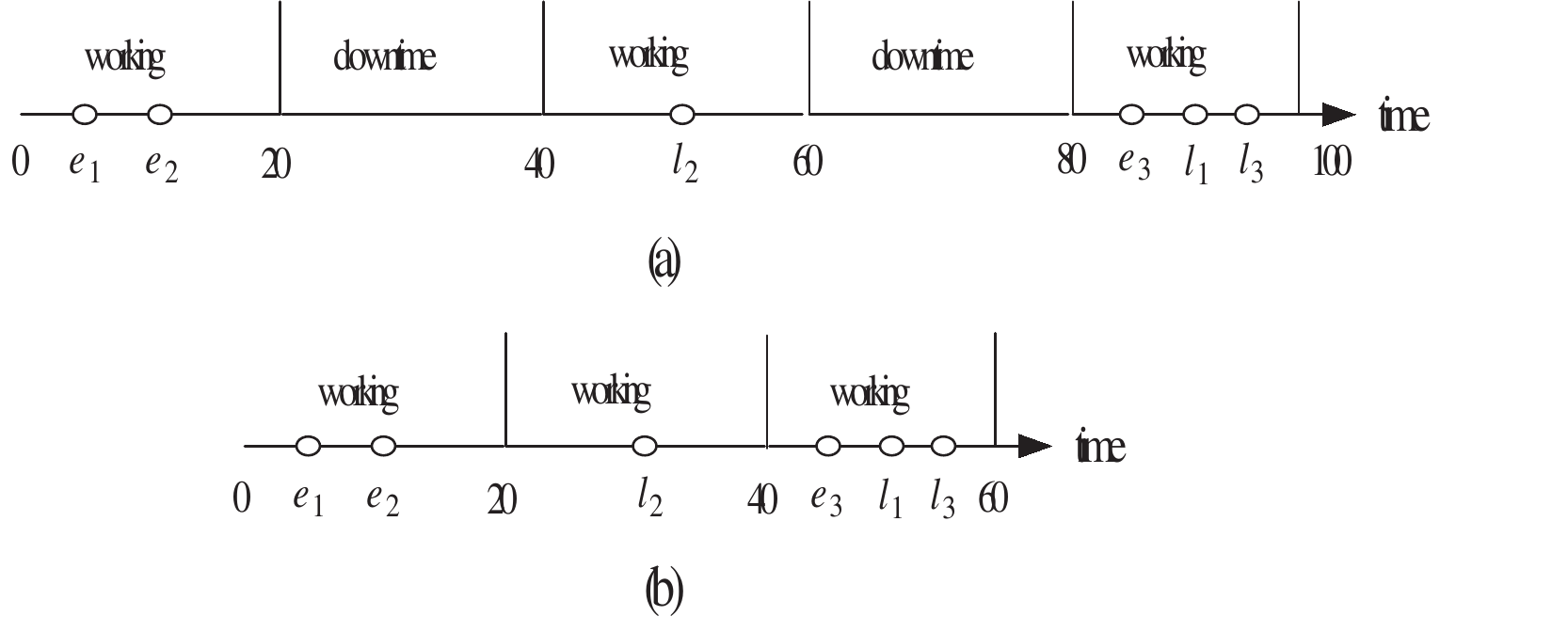}}
\end{center}
\caption{(a) The original case. (b) The case after transformation.}  \label{fig:transformation}
\end{figure}

In reality, $a_p$ should be larger than $b_{p-1}$, and the duration between $a_{p}$ and $b_{p-1}$ is the downtime for rest and recuperation. Without loss of generality, we can assume that the length of the downtime is extremely small by setting $b_{p-1} = a_p$ and imposing a break at time $b_{p-1}$. As illustrated in Figure \ref{fig:transformation}, we can easily transform the non-zero downtime cases to zero downtime ones. In Figure \ref{fig:transformation}(a), $T=20$ and the time windows of suppliers 1, 2, and 3 are $[5, 90]$, $[10, 50]$ and $[85, 95]$, respectively. After transformation, their time windows become [5, 50], [10, 30] and [45, 55] (see Figure \ref{fig:transformation}(b)).

\begin{figure}[!h]
\begin{center}
\resizebox{8cm}{!}{\includegraphics{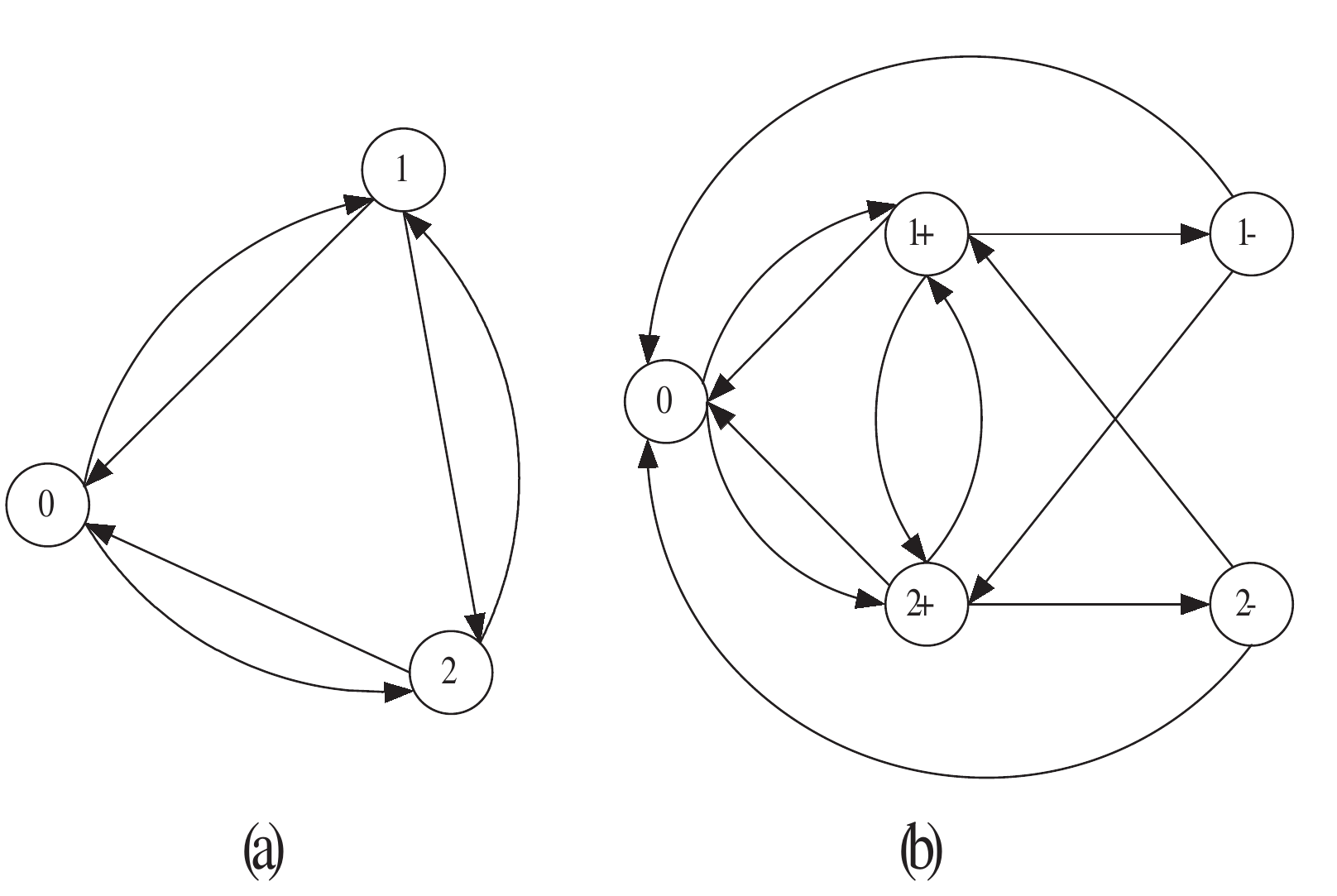}}
\end{center}
\caption{(a) $G = (V, E)$. (b) $G' = (V', E')$.} \label{fig:conversion}
\end{figure}

To further describe the MPISP, we convert the graph $G = (V, E)$ into a directed (not complete) graph $G' = (V', E')$ by the following two steps: (1) split each vertex $i\in V_C$ into two vertices $i+$ and $i-$, and create an edge $(i+, i-)$, where vertex $i+$ represents the arrival of vertex $i$ and vertex $i-$ represents the completion of supplier $i$'s service; (2) create edges $(0, i+)$, $(i+, 0)$, $(i-, 0)$, $(i+, j+)$ and $(i-, j+)$, where $i, j \in V_C$ and $i \neq j$; and (3) set $t_{i+, i-} = s_i$, $t_{0, i+} = t_{0,i}$, $t_{i+, 0} = t_{i-, 0} = t_{i, 0}$ and $t_{i+, j+} = t_{i-, j+} = t_{i, j}$. An example to illustrate this conversion is shown in Figure \ref{fig:conversion}, where Figure \ref{fig:conversion}(b) is the resultant graph derived from Figure \ref{fig:conversion}(a).

\begin{figure}[!h]
\begin{center}
\resizebox{12cm}{!}{\includegraphics{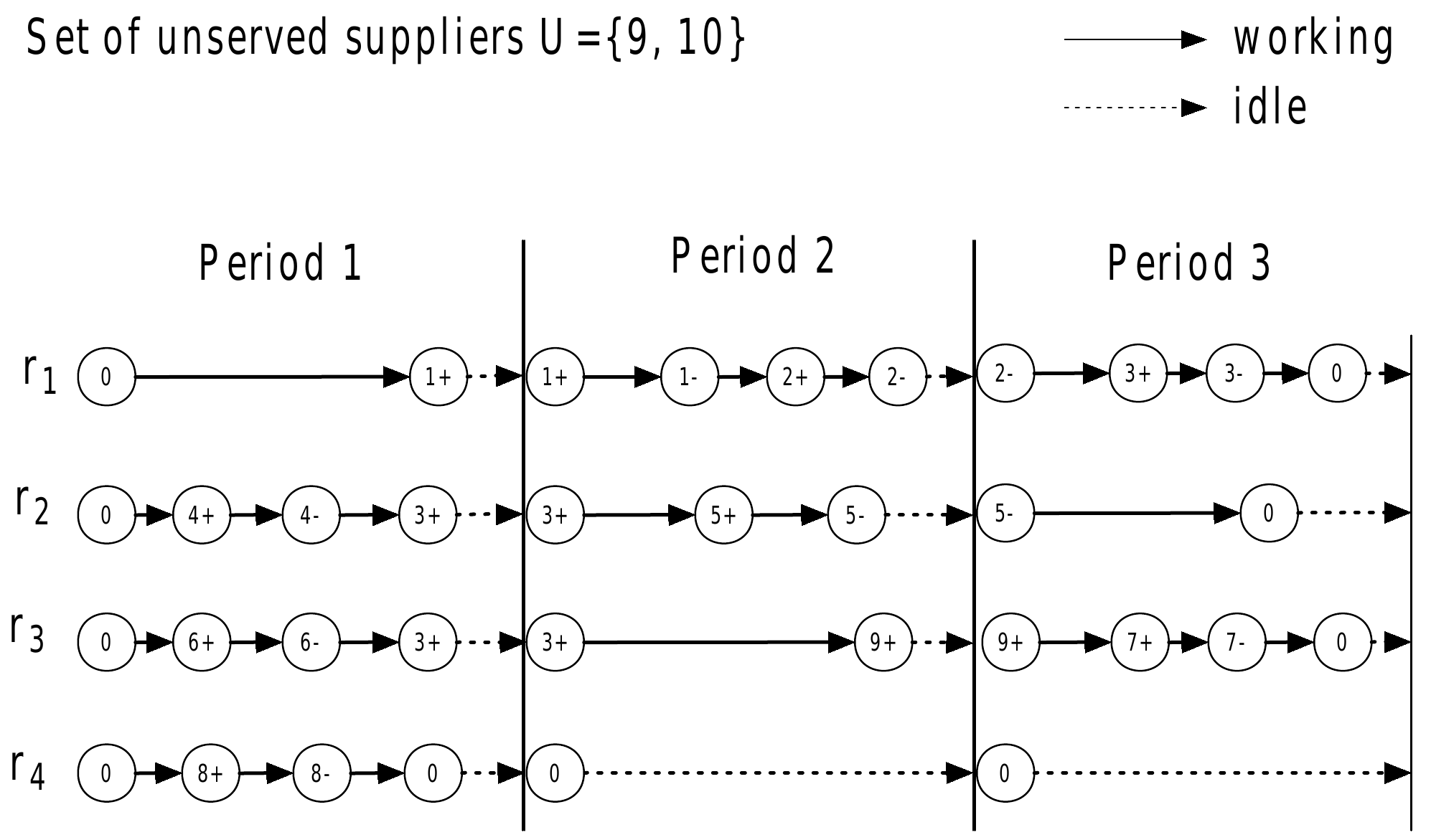}}
\end{center}
\caption{An example feasible solution to an MPISP instance.} \label{fig:example}
\end{figure}

We can denote a feasible solution of the MPISP by $S$, consisting of $m$ routes, namely $S = \{r_1, r_2, \ldots, r_m\}$. A route $r_k$ $(1\leq k \leq m)$ is divided into $w$ sub-routes by periods and therefore can be expressed as $r_k = (r_k^1, r_k^2, \ldots, r_k^w)$, where $r_k^p$ $(1 \leq p \leq w)$ denotes the trip in period $p$. If an inspector returns to the depot before period $w$, he/she will stay at the depot for the remaining periods. The sub-route $r_k^p$ is a sequence of vertices, where its starting and ending vertices are denoted by $v_s(r_k^p)$ and $v_e(r_k^p)$, respectively. If an inspector $k$ stays at the depot during the whole period $p$, we set $r_k^p = (0)$ and $v_s(r_k^p)=v_e(r_k^p) = 0$. According to the definition of our problem, an inspector must stay at vertex $v_e(r_k^p)$ for rest and will start the next trip from this vertex in period $p+1$, i.e, $v_e(r_k^p) = v_s(r_k^{p+1})$ for all $1 \leq p \leq w -1$. Obviously, the starting vertex of period $1$ and the ending vertex of period $w$ for each route must be vertex $0$. Figure \ref{fig:example} gives a feasible solution to an MPISP instance involving 10 suppliers, 4 inspectors and 3 periods.

In Figure \ref{fig:example}, the solid and dash lines denote the working (i.e., traveling or providing service) and idle statuses of the inspectors, respectively. The set $U =\{9, 10\}$ indicates that suppliers 9 and 10 are not served by any inspector. The route $r_1 = (r_1^1, r_1^2, r_1^3)$ comprises three trips, i.e., $r_1^1 = (0, 1+)$, $r_1^2 = (1+, 1-, 2+, 2-)$ and $r_1^3 = (2-, 3+, 3-, 0)$. This route shows that inspector 1 arrives at vertex 1, but does not have sufficient time to complete the service for supplier 1 in the first period. Thus, he/she has to wait until the start of the second period and then provides service to supplier 1. Subsequently, inspector 1 travels to vertex 2, completes the service of supplier 2 and stays at vertex 2 for recuperation. In the third period, inspector 1 travels from vertex 2 to vertex 3, provides service for supplier 3, and finally returns to the depot. Since each supplier $i$ can be served by at most one inspector, edge $(i+, i-)$ can be included in at most one route. In route $r_2$, after completing the service of supplier 4, inspector 2 travels to vertex 5 via a waypoint, namely vertex 3. Note that any waypoint must be the ending vertex of a certain period (and also be the starting vertex of the following period) due to the rule of triangle inequality. As shown in route $r_3$, an inspector may use two or more waypoints between two consecutively served suppliers. In this route, inspector 3 visits but does not serve supplier 9, i.e., vertex 9 only acts as a waypoint. As no route traverses edge $(9+, 9-)$, supplier 9 is not served by any inspector in this solution. The route $r_4$ illustrates that an inspector may be idle during some periods; its three trips are represented by $r_4^1 = (0, 8+, 8-, 0)$, $r_4^2 = (0)$ and $r_4^3 = (0)$.

\section{Shortest transit time}
\label{sec:spp}
\begin{figure}[!h]
\begin{center}
\resizebox{10cm}{!}{\includegraphics{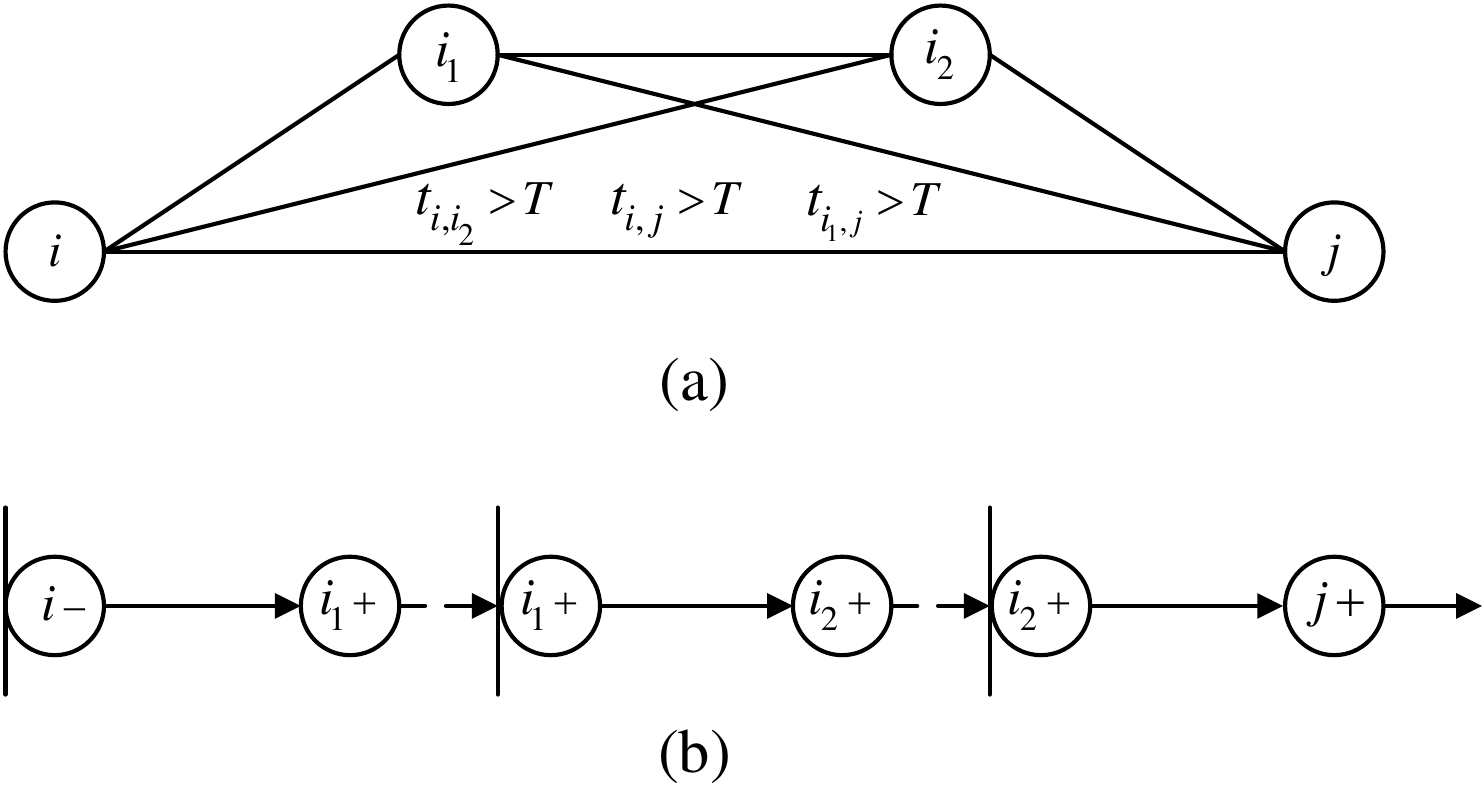}}
\end{center}
\caption{(a) The vertex locations. (b) The route associated with the shortest transit time from vertex $i$ to vertex $j$, where vertices $i_1$ and $i_2$ are waypoints.} \label{fig:transit-time}
\end{figure}

In a complete graph that satisfies the triangle inequality, the shortest path from vertex $i$ to vertex $j$ must be edge $(i, j)$. When working periods are imposed on the inspectors, edge $(i, j)$ may be unusable in some situations and therefore the shortest transit time may be greater than $t_{i, j}$. The simplest such situation can be encountered when $t_{i, j} > T$. To move from vertex $i$ to vertex $j$, an inspector has to use some waypoints and the transit time may cross several periods. We illustrate this situation in Figure \ref{fig:transit-time}, where an inspector has completed the service of supplier $i$ and departs from vertex $i$ at the beginning of a certain period.

\begin{figure}[!h]
\begin{center}
\resizebox{12cm}{!}{\includegraphics{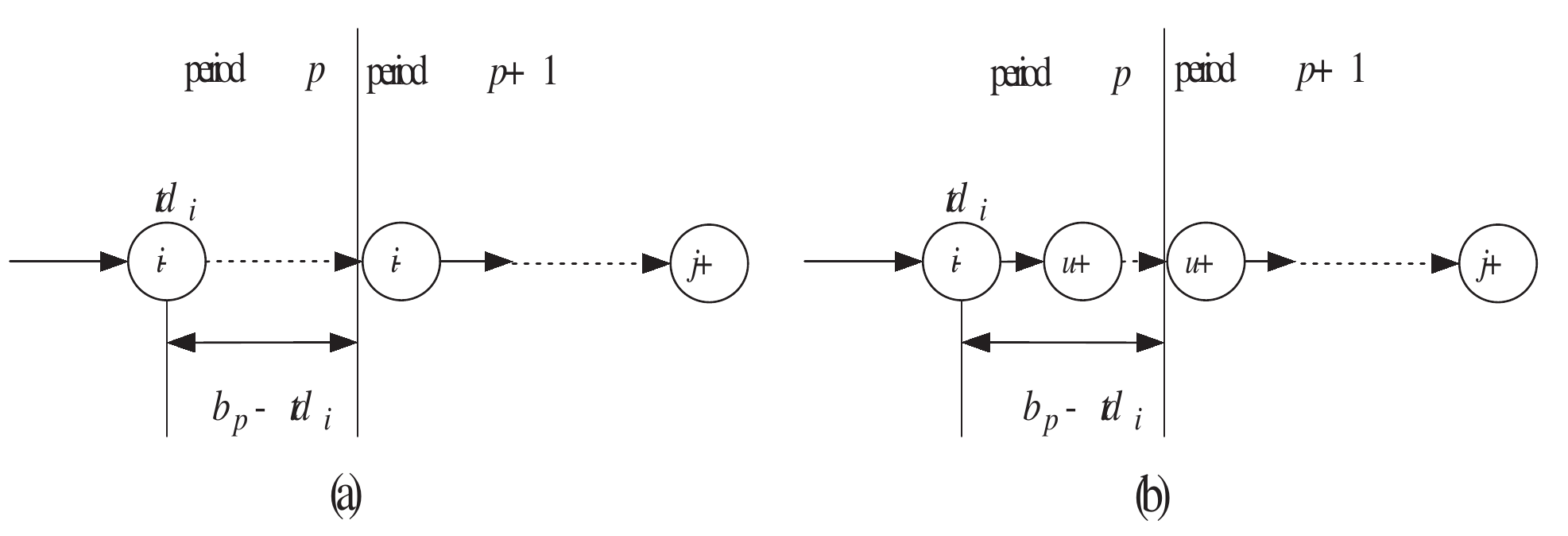}}
\end{center}
\caption{(a) Wait at vertex $i$. (b) Travel to a waypoint $u$.} \label{fig:transit-time2}
\end{figure}

Unlike the classical VRP models, in the MPISP the shortest transit times from vertex $i$ to other vertices are affected by the departure time (denoted by $dt_i$) of the inspector.  Therefore, we define $\hat{t}_{i, j} (dt_i)$ as the shortest transit time from vertex $i$ to vertex $j$ with departure time $dt_i$. If $dt_i$ is the opening time of a certain period, i.e., $dt_i=a_p$ for some $p\in P$, $\hat{t}_{i, j} (dt_i)$ can be simplified to $\hat{t}_{i, j}$. Further, we define $ceil(dt_i)$ as the closing time of the period within which $dt_i$ lies, i.e.,  if $a_p < dt_i \leq b_p$, then $ceil(dt_i) = b_p$. If $ceil(dt_i) - dt_i \geq t_{i, j}$, an inspector can travel across edge $(i, j)$ within the current period and thus $\hat{t}_{i, j} (dt_i) = t_{i, j}$. Otherwise, the inspector has to either wait at vertex $i$ until the start of the next period or travel to some waypoint $u$. We illustrate these situations in Figure \ref{fig:transit-time2}, where an inspector may travel from vertex $i$ to vertex $j$ via some waypoint.

As previously mentioned, a waypoint $u$ can only be positioned as the last or the first vertex in the trip of some period. More precisely, if an inspector travels to a waypoint $u$, he must stay at $u$ for downtime (see Figure \ref{fig:transit-time2}(b)). Taking $N(dt_i) = \{u \in V| t_{i, u} \leq ceil(dt_i) - dt_i\}$ to be the set of all vertices that can act as waypoints for vertex $i$, the value of $\hat{t}_{i,j}(dt_i)$ can be calculated by:
 \begin{align}
\hat{t}_{i, j}(dt_i) = \left \{\begin{array}{ll}
t_{i, j}, & \textrm{if $ceil(dt_i) - dt_i \geq t_{i, j}$}; \\
ceil(dt_i) - {dt}_i + \min_{u \in N(dt_i) \cup \{i\}}\{\hat{t}_{u, j}\}, & \textrm{otherwise.}
\end{array} \right. \label{exp:1}
\end{align}
The above expression shows that computing any $\hat{t}_{i, j}(dt_i)$ requires $O(n)$ time given the values of all $\hat{t}_{i, j}$, which can be calculated prior to applying any algorithm to the problem. If the last waypoint between vertex $i$ and vertex $j$ is vertex $u$, the corresponding shortest transit time, denoted by $\hat{t}^u_{i, j}$, can be obtained by:
 \begin{align}
\hat{t}^u_{i,j } = \left \{\begin{array}{ll}
\hat{t}_{i, u} + t_{u, j}, & \textrm{if $ ceil(\hat{t}_{i, u}) - \hat{t}_{i, u} \geq t_{u, j}$}; \\
ceil(\hat{t}_{i, u}) + t_{u, j}, &  \textrm{if $ ceil(\hat{t}_{i, u}) - \hat{t}_{i, u} < t_{u, j} \leq T$; } \\
+\infty, &\textrm{otherwise.}
\end{array} \right. \label{exp:extension}
\end{align}
Obviously, we have:
\begin{align}
\hat{t}_{i, j} = \min_{u \in V}\{\hat{t}^u_{i, j}\} \nonumber
\end{align}
To compute all $\hat{t}_{i, j}$, we can apply an algorithm modified from the Dijkstra's algorithm \citep{Ahuja1993}, one of the most well-known label-setting algorithms for the classical shortest path problem. This modified Dijkstra's algorithm employs expression (\ref{exp:extension}) as the extension function and has a time complexity of $O(n^2)$. Since we need to compute the shortest transit time between each vertex pair, the total time complexity for all $\hat{t}_{i, j}$ is bounded by $O(n^3)$.

\begin{figure}[!h]
\begin{center}
\resizebox{12cm}{!}{\includegraphics{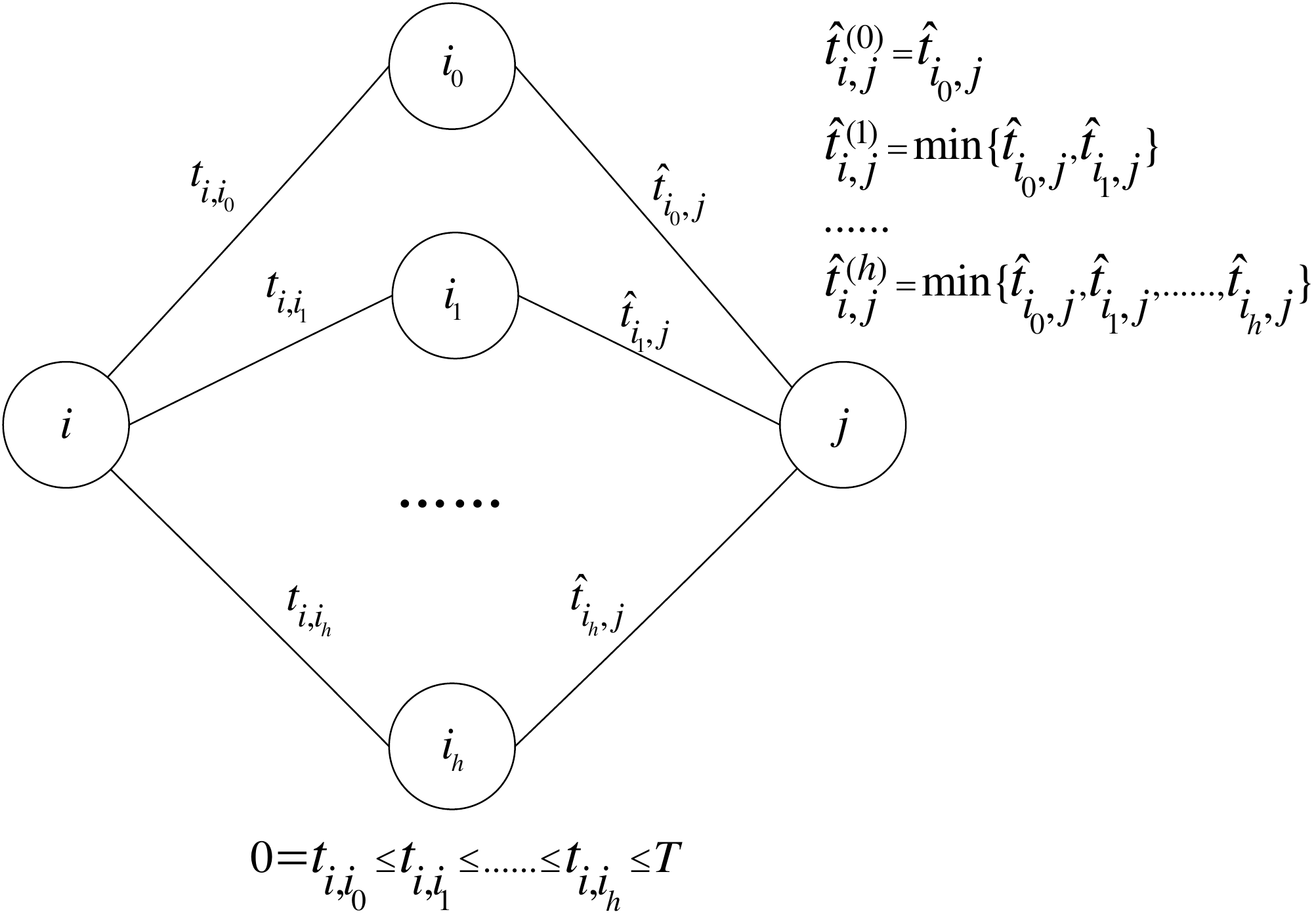}}
\end{center}
\caption{The process of computing all $\hat{t}^{(k)}_{i,j}$.} \label{fig:transit-time3}
\end{figure}

We can accelerate the computation of $\hat{t}_{i, j} (dt_i)$ by the following procedure. We first remove from the graph all edges whose lengths are greater than $T$ and then sort all neighbors $u$ of vertex $i$ in ascending order of $t_{i, u}$, generating a vertex sequence $(i_0, i_1, \ldots, i_{h})$. Note that we have $i_0 = i$ since $t_{i,i}=0$. For $0\leq k\leq h$, let $\hat{t}_{i,j}^{(k)} = \min_{0\leq k' \leq k }\{\hat{t}_{i_{k'}, j}\}$ be the shortest transit time from one of the first $k+1$ vertices in the sequence to vertex $j$. The values of all $\hat{t}_{i,j}^{(k)}$ can be computed by Algorithm \ref{alg:dp1} in time complexity of $O(n^3)$. Figure \ref{fig:transit-time3} pictorially shows the process of computing all $\hat{t}_{i,j}^{(k)}$. According to expression (\ref{exp:1}), $\hat{t}_{i,j}(dt_i)$ = $ceil(dt_i) - {dt}_i + \min_{u \in N(dt_i)\cup \{i\}}\{\hat{t}_{u, j}\}$ if $ceil(dt_i) - dt_i < t_{i,j}$. To achieve this $\hat{t}_{i,j}(dt_i)$, we identify the largest $k$ satisfying $ceil(dt_i) - dt_i \geq t_{i, i_k}$ using binary search on $t_{i,i_0}, t_{i,i_1}, \ldots, t_{i,i_h}$, and retrieve the value of $\hat{t}_{i,j}^{(k)}$, which is equal to $\min_{u \in N(dt_i)\cup \{i\}}\{\hat{t}_{u, j}\}$. The above procedure shows that the time complexity of computing $\hat{t}_{i, j}(dt_i)$ can be reduced to $O(log n)$ given that all $\hat{t}_{i,j}^{(k)}$ are available.

\begin{algorithm}[!htp]
\caption{The algorithm for preprocessing all $\hat{t}_{i,j}^{(k)}$.}
\label{alg:dp1}
\begin{algorithmic}[1]
\STATE INPUTS: all $\hat{t}_{i, j}$;
\FOR{$i= 0$ to $n$}
\STATE Sort all neighbors $u$ of vertex $i$ in ascending order of $t_{i, u}$ to generate a vertex sequence $(i=i_0, i_1, \ldots, i_h)$;
\FOR{$j = 0$ to $n$}
\STATE $\hat{t}_{i,j}^{(0)} = \hat{t}_{i, j}$;
\FOR{$k = 1$ to $h$}
\STATE $\hat{t}_{i,j}^{(k)} = \min\{\hat{t}_{i,j}^{(k-1)}, \hat{t}_{i_k, j}\}$;
\ENDFOR
\ENDFOR
\ENDFOR
\end{algorithmic}
\end{algorithm}

The computation of all $\hat{t}_{i, j}$ and $\hat{t}_{i,j}^{(k)}$ can be done in a preprocessing stage, which requires a time complexity of $O(n^3)$. The approach described in the following section needs to frequently compute $\hat{t}_{i, j}(dt_i)$. Thus, this preprocessing stage is particularly useful to save the overall computation time.

\section{Tabu search algorithm}
\label{sec:aits}
Tabu search algorithm has been successfully applied to a wide variety of routing and scheduling problems, such as the classical VRP \citep{Gendreau1994,Toth2003}, the VRPTW \citep{Chiang1997,Gordeau2001}, the three-dimensional loading capacitated VRP \citep{Zhu2012}, the job-shop scheduling problem \citep{barnes1995solving} and the nurse rostering problem \citep{burke2003tabu}. Basically, tabu search algorithm starts from an initial solution and iteratively proceeds from the {\em incumbent} solution to its best allowable neighbor. The neighborhood of a solution is a set of solutions that can be reached from that solution by a certain operation. Each type of operation corresponds to a neighborhood and the procedure of identifying the best allowable neighbor in the neighborhood is called an {\em operator}. The transition from the incumbent solution to one of its neighbors is called a {\em move}.

Our tabu search algorithm employs several operations adapted from classical operations for the VRPTW, namely {\em 2-opt, Or-opt, 2-opt$^*$, Relocate} and {\em Exchange} \citep{braysy2005vehicle}, and an {\em ejection pool} \citep{Lim2007,Nagata2009}. The most noteworthy characteristic that distinguishes these adapted operations from their classical counterparts is the procedure of checking the feasibility of the modified solution. For example, after performing an operation on a VRPTW solution, we can check the feasibility of the resultant solution in $O(1)$ time (it is assumed that for each vertex the latest arrival time that does not lead to the violation of the time windows of all successive vertices has been calculated in a preprocessing step). However, for a modified MPISP solution, we may require up to $O(n log n)$ time to check its feasibility due to the re-computation of the shortest transit times associated with the affected vertices, which will be elaborated in Section \ref{subsub:so}.

The pseudocode of our tabu search algorithm is presented in Algorithm \ref{alg:ts}, which is an iterative approach that follows a four-phase framework: initialization, local search with tabu moves, ejection pool algorithm and perturbation. At the beginning of the algorithm, we generate an initial solution $S_0$ using the function {\em best\_init} (see Section \ref{sub:i}) and then initialize both the best solution $S_{best}$ and the current solution $S$ by $S_0$. In each iteration, we first invoke the local search procedure with tabu moves (function {\em local\_search}, see Section \ref{sub:lsfm}) and set $S'$ to be the best solution found by this procedure. Subsequently, we try to improve on $S'$ by an ejection pool algorithm (function {\em EPA}, see Section \ref{sub:eip}) and then update $S_{best}$ if possible. Finally, the search process is diversified by perturbing the best solution found in this iteration. The above process is repeated until the perturbation procedure (function {\em perturb}, see Section \ref{sub:p}) is consecutively performed {\em maxPerturbation} times without improving on $S_{best}$.

\begin{small}
\begin{algorithm}[!h]
\caption{Framework of the tabu search algorithm.}
\label{alg:ts}
\begin{algorithmic}[1]
\STATE $S_0$ $\leftarrow$ {\em best\_init}();
\STATE $S_{best}$ $\leftarrow$ $S_0$ and $S$ $\leftarrow$ $S_0$;
\STATE $i$ $\leftarrow$ $0$;
\WHILE{$i \leq $ {\em maxPerturbation}}
\STATE $S'$ $\leftarrow$ the best solution found by {\em local\_search}($S$);
\STATE $S'$ $\leftarrow$ $EPA(S')$
\IF{$S'$ is better than $S_{best}$}
\STATE $S_{best}$ $\leftarrow$ $S'$ and $i$ $\leftarrow$ 0;
\ELSE
\STATE $i$ $\leftarrow$ $i + 1$;
\ENDIF
\STATE $S$  $\leftarrow$ {\em perturb} ($S'$);
\ENDWHILE
\RETURN $S_{best}$.
\end{algorithmic}
\end{algorithm}
\end{small}

\subsection{Solution representation}
\label{sub:sr}
\begin{figure}[!h]
\begin{center}
\resizebox{8cm}{!}{\includegraphics{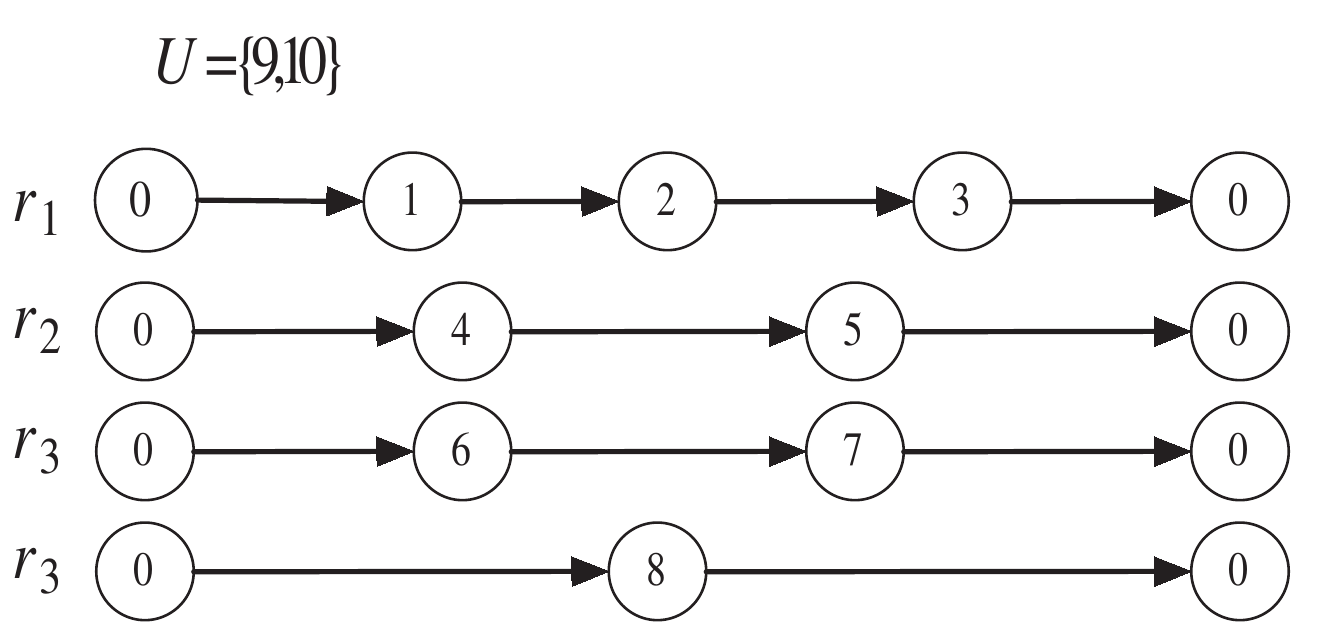}}
\end{center}
\caption{An example of the solution representation used in our tabu search algorithm.} \label{fig:solrep}
\end{figure}

In Section \ref{sec:pd}, we have used the sequences of visited vertices to represent the problem solution (see Figure \ref{fig:example}). However, in our tabu search algorithm, we represent the route of each inspector by a sequence of served suppliers. For example, Figure \ref{fig:solrep} shows a solution that is exactly the same as the one in Figure \ref{fig:example}. The routes $r_1, r_2, r_3$ and $r_4$ include the served suppliers and the ejection pool $U$ contains the leftover suppliers. All waypoints are not displayed in this solution representation and there may exist waypoints and/or breaks between two consecutively served suppliers.

\subsection{Fitness function}
\label{sub:ef}
The tabu search algorithm ranks solutions using a fitness function. It is natural to define the fitness value of a solution $S$ as the total completed workload, denoted by $P(S)$. However, many distinct solutions have the same value of $P(S)$. To further differentiate solutions, we incorporate into the fitness function another two measures denoted by $D(S)$ and $F(S)$, respectively, which is inspired by \citet{Lim2007}. As a result, the fitness function consists of three lexicographically ordered components, namely $P(S)$, $D(S)$ and $F(S)$.

The second component $D(S)$ employs a function $mv(u, S)$ that estimates the difficulty of inserting supplier $u \in U$ into the routes of solution $S$. Denoting any route in $S$ by $r = (v_0, v_1, \ldots, v_{|r|}, v_{|r|+1})$, where $|r|$ is the number of served suppliers in route $r$ and $v_0 = v_{|r|+1} = 0$, the definition of $mv(u, S)$ is given by:
\begin{align}
mv(u, S) = \min_{r\in S} mv(u, r) \nonumber
\end{align}
where
\begin{align}
&mv(u, r) = \max\Big{\{}\eta \times mv_{l}(u, r),~~ mv_{t}(u, r)\Big{\}}\label{exp:eta}\\
&mv_{l}(u, r) = \left \{\begin{array}{ll}
0, & \textrm{if $wl(r) + d_u \leq Q$}; \\
(l_0 - e_0)\times \frac{(wl(r) + d_u - Q)}{wl(r) + d_u},&  \textrm{otherwise.}
\end{array} \right. \label{exp:load}  \\
& wl(r) = \sum_{v\in r} d_v \nonumber \\
&mv_{t}(u, r) = \min_{0 \leq i \leq  |r|}c(u, v_i, r) \nonumber\\
&c(u, v_i, r) = \max \Big{\{}{ea}'_u - {l}_u, 0 \Big{\}} + \max \Big{\{} e_u - {la}'_u, 0 \Big{\}} + \max\Big{\{} {ea}'_{v_{i+1}} - {la}_{v_{i+1}}, 0 \Big{\}} \label{exp:time}
\end{align}

The cost of inserting supplier $u$ into route $r$, denoted by $mv(u, r)$, is computed based on the extent of violating the workload limit and the time-window constraint. The amount of workload is translated into time unit by expression (\ref{exp:load}), where $wl(r)$ is the cumulative workload in route $r$. If the inspector has enough capability to serve supplier $u$, namely $wl(r) + d_u \leq Q$, then no workload penalty is incurred. Otherwise, the penalty, denoted by ${mv}_{l}(u, r)$, equals the length of the depot time window multiplied by the workload violation percentage.

The penalty for time-window violation, denoted by $mv_{t}(u, r)$, considers all possible insertions. For each $v_i \in r$, we can easily find its earliest arrival time ${ea}_{v_i}$ when $(0, v_1, \ldots, v_i)$ is feasible, and its latest arrival time ${la}_{v_i}$ that does not affect the feasibility of $(v_{i+1}, v_{i+2}, \ldots, v_{|r|}, 0)$. Inserting $u$ into $r$ at the position immediately after $v_i$ creates a new route $r'$, which may be infeasible. Under the condition that $(0, v_1, \ldots, v_i)$ is feasible, we can find the earliest arrival times at $u$ and $v_{i+1}$ in $r'$, denoted by ${ea}'_u$ and  ${ea}'_{v_{i+1}}$, respectively. The partial route $(0, v_1, \ldots, v_i, u, v_{i+1})$ may be infeasible, i.e., ${ea}'_u > l_u$ and/or ${ea}'_{v_{i+1}} > l_{v_{i+1}}$. Furthermore, we can also find the latest arrival time at $u$, denoted by ${la}'_u$, that makes $(v_{i+1}, v_{i+2}, \ldots, v_{|r|}, 0)$ feasible. The penalty for time-window violation incurred by inserting $u$ between $v_i$ and $v_{i+1}$ is calculated by summing up the violations of $l_u$, $e_u$ and ${la}_{v_{i+1}}$ (see expression (\ref{exp:time})). As shown in expression (\ref{exp:eta}), the cost of inserting $u$ into $r$ takes into account both ${mv}_{l}(u, r)$ and ${mv}_{t}(u, r)$ whose relative weights are controlled by a parameter $\eta$.  After sorting the $mv(u, S)$ values of all unserved suppliers in ascending order, we can obtain a sequence $({mv}_1, \ldots, {mv}_{|U|})$, where $|U|$ is the cardinality of $U$. The value of $D(S)$ is calculated by $\sum_{i=1}^{|U|}{mv}_i/i$. We believe that the solution $S$ with smaller $D(S)$ has more chance to be improved by including the unserved suppliers.

The third component $F(S)$ is the summation of the maximal free times of all routes in $S$. The maximal free time of route $r$ is defined as $mft(r) = \max_{0 \leq i \leq |r|+1 }\{ {la}_i - {ea}_i\}$ and accordingly $F(S) = \sum_{r \in S}mft(r)$.

\subsection{Initialization}
\label{sub:i}
We obtain an initial feasible solution for the tabu search algorithm using Algorithm \ref{alg:i}. This algorithm first generates $N_{init}$ feasible solutions using the function {\em init} (see Algorithm \ref{alg:init}) and then chooses the best one as the initial solution. In each iteration of {\em init}, we begin with computing the shortest transit time ${st}^r_i$ from the tail of each route $r$ to each unserved supplier $v_i$. If $v_i$ cannot be feasibly appended at the tail of $r$, we set ${st}_i^r = + \infty$. Next, we calculate the ratio of ${st}^r_i$ to $d_i$ and set $\rho_i$ to be the minimal ratio of $v_i$ over all routes (see Algorithm \ref{alg:init}, line \ref{algorithm:init:ratio}). If the value of $\rho_i$ is positive infinity, i.e., $v_i$ cannot be appended at the tail of any route, we remove $v_i$ from $U$. Finally, we sort all suppliers in $U$ by increasing value of $\rho_i$ and relocate the $k$-th supplier $v_s$ from $U$ to the tail of the route $r$ who has $\rho_s = {st}^r_s/d_s$. The value of $k$ is a random number generated by $k = \lfloor random(0,1)^{\alpha_1} \times |U| \rfloor$, where the controlling parameter $\alpha_1 > 1$. This process is repeated until $U$ becomes empty.

\begin{small}
\begin{algorithm}[!h]
\caption{Function {\em best\_init}.}
\label{alg:i}
\begin{algorithmic}[1]
\STATE Initialize $S_{0} = \emptyset$;
\WHILE{$i \leq N_{init}$}
\STATE $S = init()$;
\IF{$S$ is better than $S_{0}$}
\STATE $S_{0}$ $\leftarrow$ $S$;
\ENDIF
\STATE $i = i+1$;
\ENDWHILE
\RETURN $S_{0}$.
\end{algorithmic}
\end{algorithm}
\end{small}

\begin{small}
\begin{algorithm}[!h]
\caption{Function {\em init}.}
\label{alg:init}
\begin{algorithmic}[1]
\STATE INPUT: the set $U$ of unserved suppliers and $m$ empty routes;
\WHILE{$U$ is not empty}
\FOR{ each $v_i$ in $U$}
\FOR{$r = 1, \ldots, m$}
\IF{$v_i$ can be feasibly appended to the tail of $r$}
\STATE ${st}_i^r$ $\leftarrow$ the shortest transit time from the last supplier of $r$ to $v_i$;
\ELSE
\STATE ${st}^r_i$ $\leftarrow$ $+\infty$;
\ENDIF
\ENDFOR
\STATE $\rho_i = \min_{r=1}^m\{{st}_i^r/d_i\}$; \label{algorithm:init:ratio}
\IF{$\rho_i = +\infty$}
\STATE Remove $v_i$ from $U$;
\ENDIF
\ENDFOR
\STATE Sort all suppliers in $U$ by increasing value of $\rho_i$;
\STATE $v_s$ $\leftarrow$ the $k$-th supplier in the sorted supplier list, where $k = \lfloor random(0,1)^{\alpha_1} \times |U| \rfloor$ and $\alpha_1 >1$;
\STATE Append $v_s$ at the tail of $r$ with ${st}_s^r/d_s = \rho_s$;
\STATE Remove $v_s$ from $U$;
\ENDWHILE
\end{algorithmic}
\end{algorithm}
\end{small}

\subsection{Local search with tabu moves}
\label{sub:lsfm}
The pseudocode of the local search procedure with tabu moves is provided in Algorithm \ref{alg:localsearch}. The following context of this subsection presents all main components of this procedure, including neighborhood structure, tabu list, aspiration and termination criteria.
\begin{small}
\begin{algorithm}[!h]
\caption{The local search procedure with tabu moves ({\em local\_search}).}
\label{alg:localsearch}
\begin{algorithmic}[1]
\STATE INPUT: the initial solution $S$;
\STATE The current best solution $S'$ $\leftarrow$ $S$ and {\em Iter} $\leftarrow$ 0;
\WHILE{$Iter \leq maxLocalIter$}
\STATE Apply the 2-opt, Or-opt, 2-opt$^*$, relocate and exchange operators on $S$ ;
\STATE $S$ $\leftarrow$ the best allowable solution found by the above operators;
\IF{$S$ is better than $S'$}
\STATE $S' \leftarrow S$ and $Iter$ $\leftarrow$ $0$;
\ELSE
\STATE $Iter$ $\leftarrow$ $Iter +1 $;
\ENDIF
\STATE Update the tabu list;
\ENDWHILE
\RETURN $S'$.
\end{algorithmic}
\end{algorithm}
\end{small}

\subsubsection{Neighborhood structure}
\label{subsub:so}
The neighborhood structure is one of the most important components that determine the size of the search space and the quality of the final solution. Our tabu search algorithm employs five neighborhood operations adapted from classical operations for the VRPTW \citep{braysy2005vehicle}, namely 2-{\em opt}, {\em Or-opt}, 2-{\em opt$^*$}, {\em Relocate} and {\em Exchange}. We treat the ejection pool as a dummy route that includes all unserved suppliers. Compared with their classical counterparts, these adapted operations require more computational efforts to check the feasibility of the resultant solution, and to update the earliest and latest arrival times at the affected suppliers.

\begin{figure}[!h]
\begin{center}
\resizebox{10cm}{!}{\includegraphics{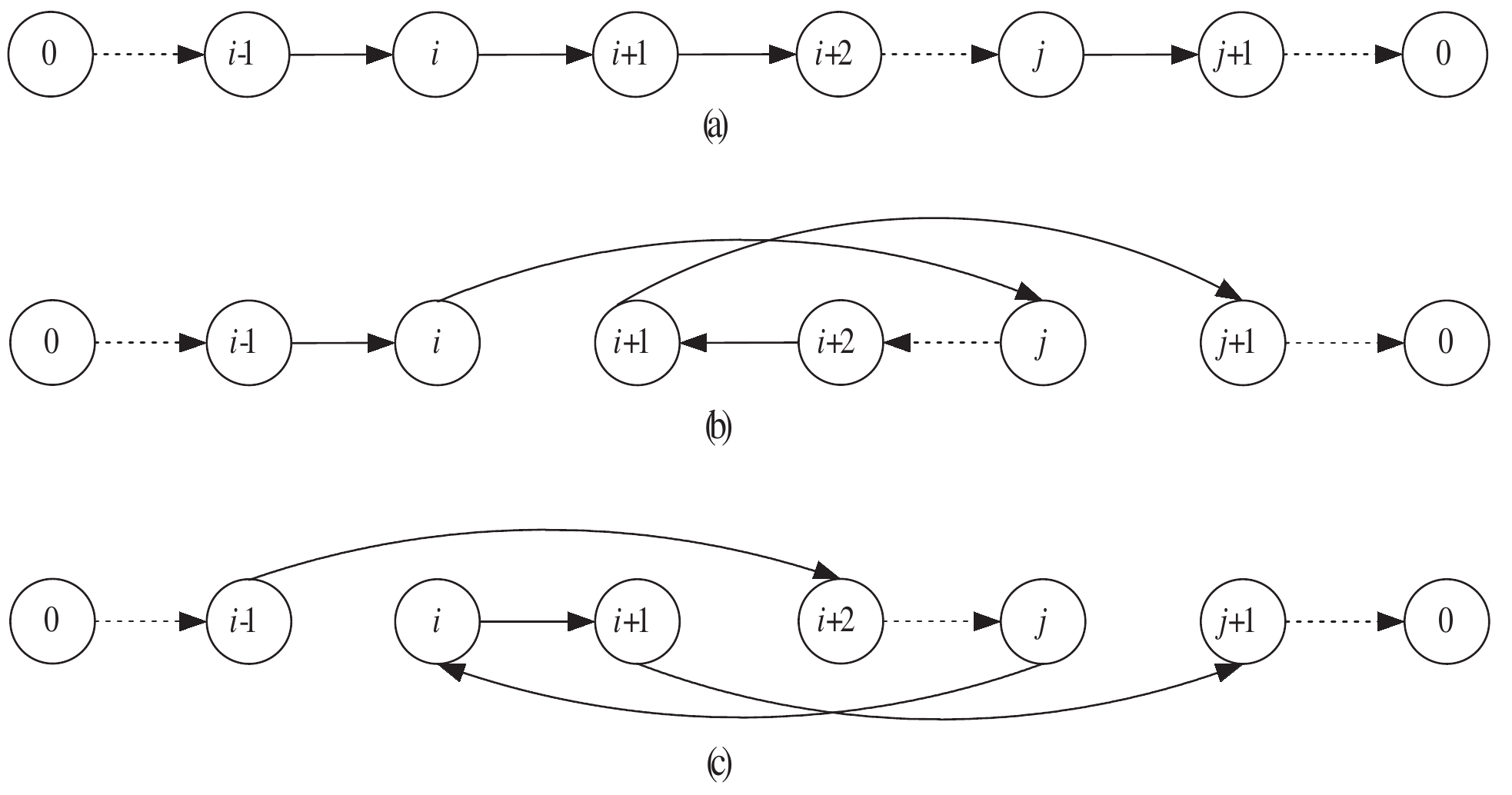}}
\end{center}
\caption{(a) The original route $r$. (b) The resultant route $r'$ after a 2-opt operation. (c) The resultant route $r'$ after an Or-opt operation.} \label{fig:2OrOpt}
\end{figure}

Figure \ref{fig:2OrOpt} illustrates the 2-opt and Or-opt operations. Assume that we are given the earliest and latest arrival times (${ea}_i$ and ${la}_i$) at each supplier $i$ in route $r$. The earliest departure time (${ed}_i$) of each supplier can be easily derived by:
 \begin{align}
{ed}_i = \left \{\begin{array}{ll}
{ea}_i + s_i, & \textrm{if $ceil(ea_i) - ea_i \geq s_{i}$}; \\
ceil({ea}_i) + {s}_i, & \textrm{otherwise.}
\end{array} \right. \nonumber
\end{align}

The 2-opt operation replaces edges ($i, i+1$) and $(j, j+1)$ with edges $(i, j)$ and $(i+1, j+1)$, and then reverses the directions of all edges between $i +1$ and $j$. The resultant route $r'$ shown in Figure \ref{fig:2OrOpt}(b) must be feasible if its subroute $(j, j - 1, \ldots, i + 1, j + 1, \ldots, 0)$ is feasible. To check the feasibility of $r'$, we need to re-calculate the earliest arrival time (${ea}'_k$) at each supplier $k$ in subroute $(j, j - 1, \ldots, i+1, j+1)$. If ${ea}'_k$ is less than $e_k$ or within $[e_k, l_k]$ for each supplier, this subroute must be feasible. If ${ea}'_{j+1}$ in $r'$ is less than or equal to ${la}_{j+1}$ in $r$, subroute $(j+1, \ldots, 0)$ must be feasible.
All $ea'_k$ can be obtained in $O(n_s log n)$ time using the procedure described in Section \ref{sec:spp}, where $n_s$ is the number of suppliers in subroute $(j, j - 1, \ldots, i + 1, j + 1)$. Therefore, it requires $O(n_slog n)$ time to check the feasibility of route $r'$. By contrast, when dealing with the VRPTW, a 2-opt operation only requires $O(n_s)$ time to accomplish the feasibility check. In addition, updating the values of ${ea}'_i$ and ${la}'_i$ for all suppliers in $r'$ requires $O(|r|log n)$ time and computing the fitness of the resultant solution requires $O(|r||U|log n)$ time.

The Or-opt operation replaces three edges ($i -1$), ($i+1, i+2$) and ($j, j+1$) with edges ($i-1, i+2$), ($j, i$) and ($i+1, j+1$); the resultant route is illustrated in Figure \ref{fig:2OrOpt}(c). After an Or-opt operation, we can also derive the time complexity for checking the feasibility of the resultant route, updating the earliest and latest arrival times at each supplier, and computing the fitness of the resultant solution in a manner similar to that used for the 2-opt operation.

\begin{figure}[!h]
\begin{center}
\resizebox{8cm}{!}{\includegraphics{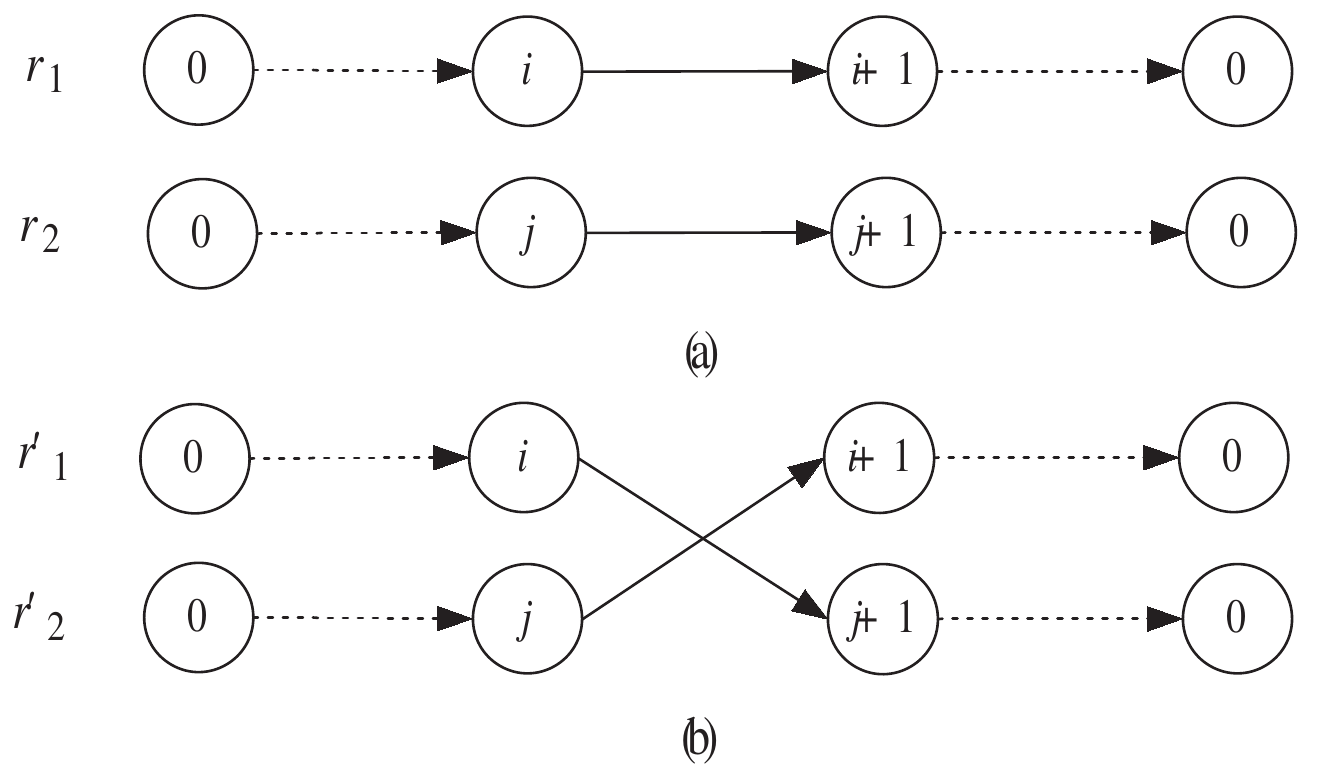}}
\end{center}
\caption{(a) The original routes $r_1$ and $r_2$. (b) The resultant routes $r'_1$ and $r'_2$ after a 2-opt$^*$ operation.} \label{fig:2Opt8}
\end{figure}

Figure \ref{fig:2Opt8} illustrates the 2-opt$^*$ operation which exchanges the latter subroutes of $r_1$ and $r_2$ by replacing edges ($i, i+1$) and ($j, j+1$) with edges ($i, j+1$) and ($j, i+1$). The feasibility of the resultant routes can be checked by simply comparing ${ea}'_{j + 1}$ (resp. ${ea}'_{i + 1}$) with ${la}_{j + 1}$ (resp. ${la}_{i + 1}$) in $O(log n)$ time. After this operation, we need $O((|r_1| + |r_2|)log n)$ time to update  ${ea}'_i$ and ${la}'_i$ in $r'_1$ and $r'_2$  and $O((|r_1| + |r_2|)|U|log n)$ time to update the fitness of the resultant solution.

\begin{figure}[!h]
\begin{center}
\resizebox{9cm}{!}{\includegraphics{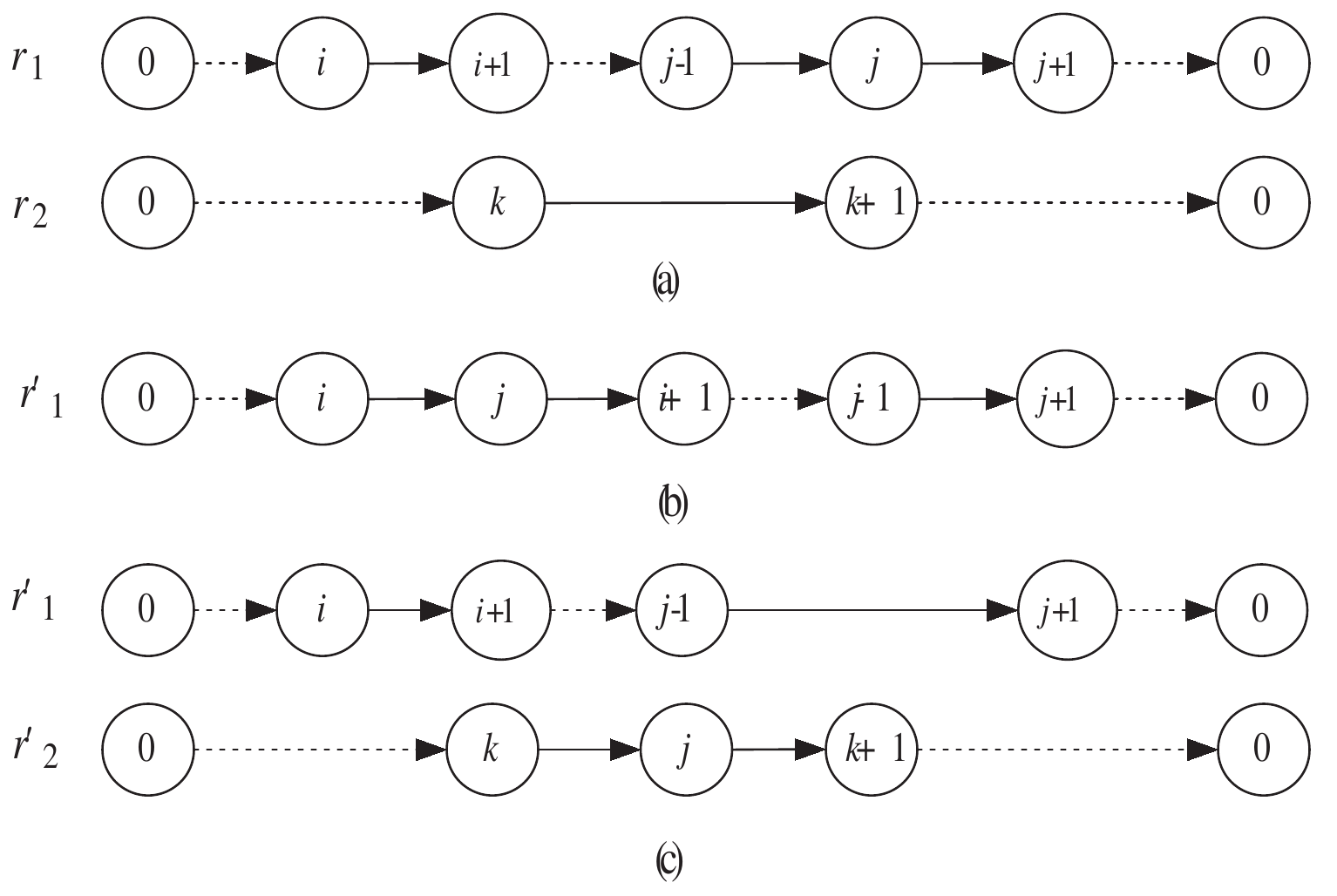}}
\end{center}
\caption{(a) The original routes $r_1$ and $r_2$. (b) The resultant routes $r'_1$ after relocating supplier $j$ between suppliers $i$ and $i + 1$. (c) The resultant routes $r'_1$ and $r'_2$ after relocating supplier $j$ between suppliers $k$ and $k + 1$.} \label{fig:relocate}
\end{figure}

The relocate operation can either relocate supplier $j$ in route $r_1$ to another position in the same route or to route $r_2$, which is illustrated in Figure \ref{fig:relocate}. In the former case, the feasibility of the resultant route shown in Figure \ref{fig:relocate}(b) can be checked by calculating ${ea}'_i$ for all suppliers in subroute $(j, i+1, \ldots, j - 1, j +1 )$. In the latter case, we only need to check the feasibility of $r'_2$ shown in Figure \ref{fig:relocate}(c), which can be done in $O(log n)$ time. The relocation operation can also relocate a supplier in the ejection pool to a certain route or vice versa.

\begin{figure}[!h]
\begin{center}
\resizebox{10cm}{!}{\includegraphics{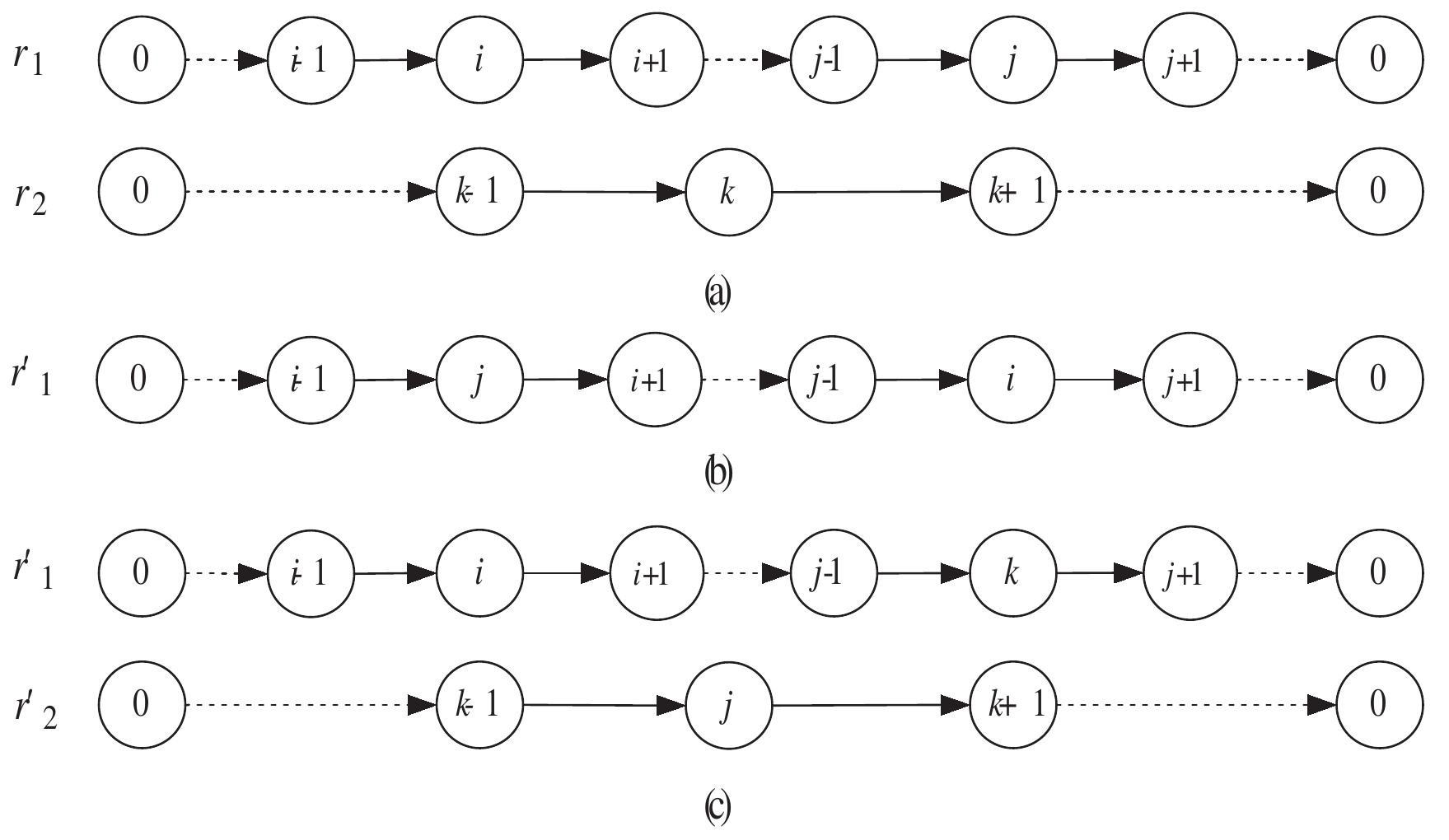}}
\end{center}
\caption{(a) The original routes $r_1$ and $r_2$. (b) The resultant routes $r'_1$ after exchanging suppliers $i$ and $j$. (c) The resultants routes $r'_1$ and $r'_2$ after exchanging suppliers $j$ and $k$.} \label{fig:exchange}
\end{figure}

The exchange operation exchanges positions of two suppliers. Figure \ref{fig:exchange}(b) shows the resultant route after exchanging the positions of two suppliers in the same route. The feasibility of this route can be checked by calculating ${ea}'_i$ for all suppliers in subroute $(j, i + 1, \ldots, j -1, i, j + 1)$. The resultant routes created by exchanging two suppliers from two different routes are shown in Figure \ref{fig:exchange}(c). The feasibility check can be done in $O(log n)$ time. This operation can also exchange a supplier in some route with a supplier in the ejection pool.

\subsubsection{Tabu list, aspiration and termination}
\label{subsub:tlm}
Tabu search algorithm employs one or more tabu lists to prevent the search process from being trapped in local optima. In our implementation, the tabu list stores edges that have been created within the previous $\xi$ iterations. A move is considered as tabu if it attempts to remove the edges in the tabu list. The tabu restriction can be overridden if the aspiration criterion is satisfied. Specifically, we allow the tabu moves to be performed if the solutions they result in are better than the current best solution $S'$. The solutions that are created by non-tabu moves or by aspiration are called {\em allowable} neighbors. All allowable moves are stored in a candidate list and sorted according to the fitness values of their resultant solutions. The best candidate is performed to generate the next incumbent solution. We terminate the local search procedure when {\em maxLocalIter} consecutive iterations are unable to improve on $S'$.

\subsection{Ejection pool}
\label{sub:eip}
Ejection pool has been previously used in the algorithms for reducing the number of routes for some routing problems (see for example \citet{Lim2007,Nagata2009,Cheang2012}). Our ejection pool algorithm (EPA) is presented in Algorithm \ref{alg:epa}. The initial solution $S'$ of this algorithm is the best solution found by function {\em local\_search}. Since $S'$ is a local optimum, no supplier in the ejection pool can be feasibly inserted into $S'$. The EPA generates a candidate solution based on $S'$ for each of the unserved suppliers by an insertion-ejection procedure. If the best candidate solution is superior to $S'$, then $S'$ is updated.

\begin{small}
\begin{algorithm}[!h]
\caption{The ejection pool algorithm ({\em EPA}).}
\label{alg:epa}
\begin{algorithmic}[1]
\STATE INPUT: the initial feasible solution $S'$;
\FOR{each $u \in U$}
\STATE Evaluate all insertion positions using the function $c(i, u, j)$;
\STATE $S_u$ $\leftarrow$ the resultant solution after inserting $u$ into $S'$ at the best position;
\STATE Eject suppliers one by one using the function $c(i)$ until $S_u$ becomes feasible;
\STATE Improve $S_u$ by function {\em local\_search} with $\xi = 0$;
\ENDFOR
\IF{the best candidate solution $S_u$ is better than $S'$}
\STATE $S'$  $\leftarrow$ $S_u$;
\ENDIF
\RETURN $S'$.
\end{algorithmic}
\end{algorithm}
\end{small}

For each $u\in U$, we evaluate its insertion positions using function $c(i, u, j)$, namely the cost of inserting $u$ between two consecutively served suppliers $i$ and $j$, which is defined as:
\begin{align}
c(i, u, j) = \beta_1\times d_u  -  \beta_2 \times \bigg{(}\max\Big{\{} 0, ea'_u - l_u \Big{\}} + \max \Big{\{} 0, ea'_j - l_j \Big{\}} \bigg{)}, \nonumber
\end{align}
where $\beta_1$ and $\beta_2$ are controlling parameters, and $ea'_u$ and $ea'_j$ are the earliest arrival times at suppliers $u$ and $j$ after inserting $u$ between $i$ and $j$. The position with the smallest value of $c(i, u, j)$ is selected for insertion.

The target route $r_t$ becomes infeasible after the insertion. Thus, some of its suppliers (except the newly inserted one) need to be ejected one by one until its feasibility is restored. The supplier $i$ to be ejected is determined based on the value of $c(i)$, which is defined as:
\begin{align}
c(i) = \beta_3 \times d_i + \beta_4 \times \max \Big{\{}wl(r'_t) - Q, 0 \Big{\}} +  \beta_5 \times violation_{tw}(r'_t), \nonumber
\end{align}
where $\beta_3$, $\beta_4$ and $\beta_5$ are controlling parameters, $r'_t$ is the resultant route after removing $i$ from $r_t$, and $violation_{tw}(r'_t)$ is the total time-window violation of all suppliers in route $r'_t$, defined as:
\begin{align}
violation_{tw}(r'_t) = \sum_{j \in r'_t} \max\Big{\{} 0, ea'_j - l_j \Big{\}}. \nonumber
\end{align}
The supplier $i$ with minimal $c(i)$ is ejected from the route $r_{t}$.

After performing the insertion-ejection procedure on each $u \in U$, we obtain $|U|$ candidate solutions, which are further improved by {\em local\_search} without tabu moves, namely $\xi = 0$. The solution $S'$ is updated by the best candidate solution if possible.

\subsection{Perturbation}
\label{sub:p}
The perturbation procedure is a diversification scheme that helps the search process escape from local optima. Our perturbation procedure (function {\em perturb}) randomly removes some suppliers from the solution $S'$ following the rule that the suppliers with smaller workloads have more chances to be removed. Given a solution $S'$, we sort the served suppliers in non-decreasing order of workloads, generating a supplier list $(v_1, v_2, \ldots, v_{n'})$. The probability of removing the $k$-th supplier is determined by:
\begin{align}
p_{min} + (p_{max} - p_{min}) \times \frac{k}{n'},	\label{eqn:prob}
\end{align}
where $p_{min}$ and $p_{max}$ are controlling parameters satisfying $0 \leq p_{min} \leq p_{max} \leq 1$. It implies that the supplier with larger workload has smaller probability to be kicked out.

The tabu search algorithm performs at least {\em maxPerturbation} iterations (see line 4 -- 13, Algorithm \ref{alg:ts}). We store the best solution $S'$ found within each iteration in a solution list. The solutions with equal values of $P(S)$, $D(S)$ and $F(S)$ are regarded as the same. The number of times that the current solution $S'$ appears in the solution list is represented by $N_{rep}$. When $N_{rep}$ grows large, we expect that the perturbed solution deviates far from the current solution $S'$. To this end, we replace $p_{min}$ and $p_{max}$ in expression (\ref{eqn:prob}) by:
\begin{align}
p_{min} \leftarrow p_{min} + p_{\Delta}\times \min\{N_{rep}, N_{max}\},	\nonumber \\
p_{max} \leftarrow p_{max} + p_{\Delta}\times \min\{N_{rep}, N_{max}\},	\nonumber
\end{align}
where $p_{\Delta}$ is a controlling parameter, and the parameter $N_{max}$ is used to set an upper bound for the probability. The introduction of $N_{max}$ can help avoid the overly large probability, which would cause the process to degenerate into an ineffective multi-start method.

\section{Upper bound}
\label{sec:ub}
The solutions generated by our tabu search algorithm are lower bounds to the MPISP. In this section, we construct a constrained knapsack model to produce an upper bound for the MPISP; this model is motivated by \citet{Lau2003}.

The time windows of all suppliers can be adjusted based on the following straightforward observations. For a supplier $i$, if the opening time of its time window lies within period $p$, i.e., $a_p \leq e_i \leq b_p$, and it is impossible to complete its service during that period, i.e., $b_p - e_i < s_i$, then the real earliest service starting time for supplier $i$ should be the opening time $a_{p+1}$ of the next period. Consequently, in this situation $e_i$ can be updated by $e_i = a_{p+1}$. Analogously, if $a_p \leq l_i \leq b_p$ and $b_p - l_i < s_i$, we can update $l_i$ by $l_i = b_p - s_i$. Let $\lambda_i = ceil({l}_i + s_i) + \hat{t}_{i, 0}$ denote the time of returning to the depot immediately after severing supplier $i$ with the service starting time $l_i$. We construct a supplier set $V_G =\{g_1, g_2, \ldots, g_m\} \subseteq V_C$ that contains $m$ distinct suppliers satisfying $\lambda_i \geq \lambda_j$ for all $i \in V_G$ and $j \in V_C\backslash V_G$.

Obviously, if $e_i > l_i$, supplier $i$ cannot be served by any inspector and vertex $i$ can only be used as a waypoint. We define $f_1(i) = 0 $ if $e_i > l_i$ and $f_1(i) = 1$ otherwise. Suppose $e_i$ and $l_i$ lie within periods $p_1$ and $p_2$, respectively. We define $f_2(i, p)= 1$ if $p_1 \leq  p \leq p_2$ and $f_2(i, p)= 0$ otherwise, where $f_2(i, p) =1$ indicates that supplier $i$ could probably be served during period $p$. Furthermore, we use $f_3(i, j) =1$ to indicate that it is possible for an inspector to serve both suppliers $i$ and $j$ when supplier time windows, workload capacity and working periods are not considered. Thus, the definition of $f_3(i, j)$ is:
\begin{displaymath}
f_3(i, j) = \left \{\begin{array}{ll}
1, & \textrm{if $e_i + s_i + t_{i,j} \leq l_j \textrm{ or } e_j + s_j + t_{j, i} \leq l_i$}; \\
0, & \textrm{otherwise.}
\end{array} \right.
\end{displaymath}
We denote by $r_{i}^p$ the time required to directly travel from supplier $i$ to its nearest neighbor who could probably be served by the same inspector during period $p$. Define set $V_{i}^p = \{j \in V| j \neq i, f_2(j, p)=1, f_3(i, j)=1 \}$. We set $r_{i}^p = +\infty$ if $V_{i}^p$ is empty, and otherwise $r_{i}^p = \min_{j \in V_i}\{t_{i, j}\}$. Let $x_{i, k, p}$ be a binary decision variable that equals 1 if supplier $i$ is served by inspector $k$ during period $p$, and 0 otherwise. The optimal solution value of the following integer programming model gives an upper bound to the MPISP:
\begin{small}
 \begin{align}
& \max\sum_{i \in V_C} \sum_{k \in K}\sum_{p \in P}d_ix_{i, k, p} \label{eqn:up:obj} \\
\mbox{s.t.}~~&\sum_{k \in K}\sum_{p\in P} x_{i, k, p} \leq f_1(i), ~\forall~i \in V_C\label{eqn:pp:c1}\\
&\sum_{k \in K}x_{i, k, p} \leq f_2(i, p), ~\forall~ i \in V_C, ~p \in P\label{eqn:pp:c2}\\
& \sum_{p \in P} (x_{i, k, p} + x_{j, k, p}) \leq 1 + f_3(i, j), ~\forall i, j \in V_C, i\neq j, k \in K \label{eqn:pp:c3}\\
& \sum_{i \in V_C}\sum_{p \in P}d_ix_{i, k, p} \leq Q, ~\forall~k \in K \label{eqn:pp:c4}\\
&\sum_{i \in V_C}\sum_{p \in P}\Big{(}s_i + \min \{r_{i}^p, r_{i}^{p+1}, \ldots, r_{i}^w\}\Big{)}x_{i, k, p} + r_{0}^1\leq \lambda_{g_k}, ~\forall~ k \in K \label{eqn:pp:c5}\\
&\sum_{i \in V_C}x_{i, k, p}(s_i + r_{i}^p) - \max_{i \in V_C}r_{i}^p \leq T, ~\forall~k \in K, ~p \in P \label{eqn:pp:c6}\\
&x_{i, k, p} \in \{0, 1\}, ~\forall~i \in V_C, ~k \in K, ~p \in P \label{eqn:pp:c7}
\end{align}
\end{small}
The objective (\ref{eqn:up:obj}) is to maximize the total completed workload. Constraints (\ref{eqn:pp:c1}) state that each supplier must be assigned to at most one inspector and be served in at most one period. Constraints (\ref{eqn:pp:c2}) guarantee that if supplier $i$ cannot be served by any inspector during period $p$, all relative variables must be set to zero. Constraints (\ref{eqn:pp:c3}) ensure that if $f_3(i, j) = 0$, suppliers $i$ and $j$ cannot be served by the same inspector. As the inspector workload capacity cannot be exceeded, Constraints (\ref{eqn:pp:c4}) apply. When supplier $i$ is served by some inspector, the time it consumes must be at least the sum of $s_i$ and the traveling time to its nearest neighbor. Any feasible solution to the MPISP must have $m$ routes, each of which has an earliest time of returning to the depot. It is easy to observe that the $k$-th largest return time must be less than or equal to the $k$-th largest $\lambda_{g_k}$. Obviously, the sum of $s_i + \min \{r_{i}^p, r_{i}^{p+1}, \ldots, r_{i}^w\}$ associated with all suppliers covered by a route should be less than or equal to the length of that route, which is capped by $\lambda_{g_k}$. Therefore, Constraints (\ref{eqn:pp:c5}) hold. After completing the service of supplier $i$, the inspector may stay at vertex $i$ until the start of the next period. The difference between the total $s_i + r_{i}^p$ of all suppliers served in each period and the largest $r_{i}^{p}$ must be less than or equal to the period length, which is ensured by Constraints (\ref{eqn:pp:c6}).

Any feasible solution to the MPISP must be also feasible to the constrained knapsack model (\ref{eqn:up:obj}) -- (\ref{eqn:pp:c7}). The knapsack problem and many of its variants have been well-studied and can be efficiently handled by several commercial mathematical programming solvers.

\section{Computational experiments}
\label{sec:ce}
Our tabu search (TS) algorithm was coded in C++ and compiled using the gcc 4.6.1 compiler, and was tested on a Dell server with an Intel Xeon E5430 2.66GHz CPU, 8 GB RAM and running Linux-CentOS-5.0 64-bit operating system. The algorithm was configured with determined parameter settings: $\eta = 1.0$, $N_{init} = 100$, $\alpha_1 = 5$, {\em maxPerturbation = 4}, {\em maxLocalIter = 200}, $\xi = 100$, $\beta_1 = 0.6$, $\beta_2 =\beta_3 = \beta_4 = 0.4$, $\beta_5 = 0.2$, $p_{min}=0.05$, $p_{max}= 0.30$, $p_{\Delta}= 0.1$ and $N_{max}= 5$. The MPISP reduces to the traditional TOPTW when $w = 1$ and $Q = + \infty$. In our experiments, we first applied the TS algorithm to the TOPTW instances and compared the results with the recent results reported in \citet{vansteenwegen2009iterated,Lin2012Simulated,Labadie2012Team,Hu2014Iterative}. Next, we conducted experiments on the MPISP instances generated from the Solomon's VRPTW instances \citep{Solomon1987}. Since our TS algorithm is not deterministic, we solved each instance ten times. Finally, we achieved an upper bound for each MPISP instance by solving the model (\ref{eqn:up:obj}) -- (\ref{eqn:pp:c7}) using ILOG CPLEX 12.1 with default settings. Computation times reported are in CPU seconds on this server. All instances and detailed results can be found online at: \url{http://www.computational-logistics.org/orlib/mpisp/}.

\subsection{Test instances}
\label{sub:ig}
We considered the TOPTW instances used in \citet{vansteenwegen2009iterated,Hu2014Iterative}, which can be accessed at \url{http://www.mech.kuleuven.be/en/cib/op}. \citet{Hu2014Iterative} classified these instances into two categories, namely ``INST-M'' and ``OPT''; the instances in category INST-M have unknown optimal solution values while the optimal solution values of the instances in category OPT are given.

The TOPTW instances in category INST-M were constructed by \citet{Montemanni2009An} based on the OPTW instances by considering the number of vehicles taken from set \{1, 2, 3, 4\}. These OPTW instances were designed by \citet{Righini2009Decremental} using 56 Solomon's VRPTW instances \citep{Solomon1987} and 20 Cordeau's multi-depot VRP (MDVRP) instances \citep{cordeau1997tabu}. The Solomon's VRPTW instances, each containing 100 customers, are divided into six groups, namely C1 (c101 -- c109), C2 (c201 -- c208), R1 (r101 -- r112), R2 (r201 -- r211), RC1 (rc101 -- rc108) and RC2 (rc201 -- rc208). The numbers of customers in the MDVRP instances (pr01 -- pr20) range from 48 to 288. The TOPTW instances were obtained from the VRPTW or MDVRP instances by the following two steps: (1) set the profit collected at each customer to be the demand of this customer, and (2) remove the vehicle capacity restriction. The instances in category OPT are the same as the instances in category INST-M except for the number of vehicles available. \citet{vansteenwegen2009iterated} designed the instances in category OPT by setting the number of vehicles in each aforementioned TOPTW instance except for pr11 -- pr20 to the number of vehicles appearing in the solution of the corresponding VRPTW instance. This implies that with such number of vehicles, all customers can be visited and the optimal objective value must be equal to the total profits of all customers. Therefore, we have 76 $\times$ 4 = 304 instances in category INST-M and 66 instances in category OPT, for a total of 370 TOPTW instances.

We generated 12 MPISP instances from each Solomon's VRPTW instance by taking the values of $w$ and $m$ from \{1, 3, 5\} and \{7, 9, 11, 13\}, respectively, for a total of 72 instance groups (each instance group is identified by the name of Solomon's instance group, $w$ and $m$) and 672 instances. The workload of each supplier is set to be the demand of the corresponding vertex. The duration of each period is set to $T = (l_0 - e_0)/ w$ and the workload limit of each inspector is set to 200. The total profits in the MPISP instances related to the Solomon's instance groups C1, C2, R1, R2, RC1 and RC2 are 1810, 1810, 1458, 1458, 1724 and 1724, respectively.

\subsection{Results for the TOPTW instances}
\label{sub:ra}
To evaluate the performance of our algorithm based on the TOPTW instances, we considered the following four state-of-the-art existing algorithms in our comparisons:
\begin{itemize}
\item ILS: the iterated local search algorithm by \citet{vansteenwegen2009iterated}.
\item SSA: the slow version of the simulated annealing algorithm by \citet{Lin2012Simulated}.
\item GVNS: the LP-based granular variable neighborhood search algorithm by \citet{Labadie2012Team}.
\item I3CH: the iterative three-component heuristic (I3CH) by \citet{Hu2014Iterative}.
\end{itemize}
\citet{Lin2012Simulated} proposed two versions of simulated annealing algorithm for the TOPTW, namely a fast version and a slow one. Compared with the fast version, the slow simulated annealing algorithm (SSA) is able to find better solutions at the expense of more computation time. As we are more concerned with solution quality, we used the SSA rather than the fast version in the comparisons. For each TOPTW instance, the ILS, SSA and I3CH were executed only once while the GVNS was performed five times. Although these algorithms were coded in different programming languages and executed on different computational environments (see Table \ref{tab:ev}), we believe that there is no dramatic difference between the speeds of these algorithms and it is acceptable to directly compare their computation times.

\begin{table*}[ht]
\normalsize{\caption{Programming languages and experimental environments.} \label{tab:ev}}
\begin{center}
\scalebox{0.85}{
\begin{tabular}{ccc}
\hline
 Algorithm &  Language  &  Experimental environment  \\
\hline
        ILS & N/A & Intel Core 2 2.5 GHz CPU, 3.45 GB RAM\\
         SSA &     C &    Intel Core 2 2.5 GHz CPU \\
         GVNS &     Embarcadero Delphi 2010 &    Intel Pentium (R) IV 3 GHz CPU  \\
             I3CH &Java   &     Intel Xeon E5430 CPU clocked at 2.66 GHz, 8 GB RAM  \\
                    TS & C++   &      Intel Xeon E5430 CPU clocked at 2.66 GHz, 8 GB RAM  \\
\hline
\end{tabular}
}
\end{center}
\end{table*}


Table \ref{tab:stat} summarizes the results of the TS, ILS, SSA, GVNS and I3CH for the TOPTW instances. We first identified the best solution value (BSV) obtained by these five algorithms and then computed the ratio of the best solution value produced by each algorithm to the BSV.  The columns ``Ratio (\%)'' and ``Time (s)'' show the average ratios and average computation times of all instance groups. Since \citet{Labadie2012Team} did not report the best solution value in their article, we filled the corresponding cells with ``N/A'' and ignored these cells when calculating the overall average ratio.  Beside the name of each algorithm, we give the number of times it was run for each instance. The overall average values of ``Ratio (\%)'' and ``Time (s)'' are presented in the last row and the best ratios in each row are marked in bold. All of the detailed solutions can be found in Appendix B.

\begin{table*}[htp]
\normalsize{\caption{The summarized results for the TOPTW instances.} \label{tab:stat}}
\begin{center}
\scalebox{0.5}{
\begin{tabular}{ccccccccccccccc}
\hline
\multirow{2}{*}{Instance Group} & \multicolumn{ 2}{c}{ILS (1 run)} &            & \multicolumn{ 2}{c}{SSA (1 run)} &            &    \multicolumn{ 2}{c}{GVNS (5 runs)} &            & \multicolumn{ 2}{c}{I3CH (1 run)} &            & \multicolumn{ 2}{c}{TS (10 runs)} \\
\cline{2 -3} \cline{5 -6} \cline{8 - 9} \cline{11 - 12} \cline{14 - 15}
 &  Ratio (\%) &   Time (s) &            &  Ratio (\%) &   Time (s) &            &  Ratio (\%)  &   Time (s) &            &  Ratio (\%) &   Time (s) &            &  Ratio (\%)  &   Time (s) \\
\hline
       $m$ =1 &            &            &            &            &            &            &                     &            &            &            &            &                        &            &            \\

        C1 &    0.9889  &       0.3  &            &    {\bf 1.0000}  &      21.1  &            &    0.9944   &     166.5  &            &   {\bf 1.0000}  &      25.2  &            &   {\bf 1.0000}    &       7.2  \\

        C2 &    0.9772  &       1.7  &            &    0.9987  &      37.5  &            &    0.9945   &     192.4  &            &    0.9960  &      84.4  &            &   {\bf 1.0000 }    &      68.2  \\

        R1 &    0.9815  &       0.2  &            &    0.9995  &      23.3  &            &    0.9834    &      29.4  &            &    0.9950  &      28.6  &            &    {\bf 1.0000}    &       4.0  \\

        R2 &    0.9731  &       1.7  &            &    0.9891  &      45.8  &            &    0.9776   &      33.8  &            &    0.9916  &     176.2  &            &   {\bf  0.9970 }   &     110.7  \\

       RC1 &    0.9708  &       0.2  &            &   {\bf  1.0000}  &      22.2  &            &    0.9812    &       9.8  &            &    0.9834  &      25.5  &            &    0.9958   &       2.9  \\

       RC2 &    0.9699  &       1.6  &            &    0.9947  &      50.3  &            &    0.9789    &      16.0  &            &    0.9774  &     119.3  &            &   {\bf  0.9974}    &     157.2  \\

pr01 -- pr20  &    0.9318  &       1.9  &            &    0.9801  &     137.3  &            &    0.9850    &      18.3  &            &    0.9768  &     119.6  &            &   {\bf  0.9938 }   &     536.8  \\

\hline
       $m$ =2 &            &            &            &            &            &            &                       &            &            &            &            &            &                   &            \\

        C1 &    0.9906  &       1.1  &     &   {\bf  1.0000} &   26.4  &            &    0.9953    &     139.5  &    &    {\bf 1.0000}  &      87.0  &            &    {\bf 1.0000}   &      21.8  \\

        C2 &    0.9746  &       3.5  &            &    0.9882  &      53.7  &            &    0.9975   &      33.8  &            &    0.9933  &     401.2  &            &    {\bf  1.0000}    &      63.8  \\

        R1 &    0.9777  &       0.9  &            &    {\bf 0.9991}  &      36.6  &            &    0.9895   &      60.3  &        &    0.9955  &      63.0  &            &    0.9970   &      14.1  \\

        R2 &    0.9702  &       2.3  &            &    0.9917  &      91.4  &            &    0.9909    &      14.7  &            &    0.9955  &     526.8  &            &    {\bf 0.9992}   &      91.5  \\

       RC1 &    0.9771  &       0.7  &            &    {\bf 1.0000}  &      40.5  &            &    0.9940    &      20.3  &            &    0.9928  &      58.9  &            &    0.9952    &      11.6  \\

       RC2 &    0.9593  &       2.2  &            &    0.9883  &      80.1  &            &    0.9839   &      12.8  &            &    0.9945  &     439.7  &            &   {\bf 0.9995 }   &     243.2  \\

pr01 -- pr20  &    0.9311  &       5.0  &            &    0.9699  &     187.8  &            &    0.9915   &      60.8  &            &    0.9825  &     275.8  &            &   {\bf 0.9935 }   &   1,809.5  \\
\hline
       $m$ = 3 &            &            &            &            &            &            &                      &            &            &            &            &            &                        &            \\

        C1 &    0.9745  &       1.5  &            &    0.9967  &      35.3  &            &    0.9955    &     165.0  &            &    0.9989  &     190.2  &            &   {\bf 1.0000}    &      31.1  \\

        C2 &    0.9807  &       2.2  &            &    0.9876  &      59.7  &            &    0.9993    &       7.7  &            &    {\bf 1.0000 } &      12.3  &            &  {\bf  1.0000  }  &       6.8  \\

        R1 &    0.9832  &       1.7  &            &    0.9972  &      56.1  &            &    0.9889    &      73.9  &            &  {\bf  0.9990 }  &     118.3  &            &    0.9983   &      16.6  \\

        R2 &    0.9970  &       1.4  &            &    0.9992  &      41.9  &            &    0.9989    &       7.0  &            &    0.9999  &      90.8  &            &  {\bf  1.0000  }  &      14.8  \\

       RC1 &    0.9695  &       1.1  &            &    0.9946  &      42.8  &            &    0.9918    &      33.7  &            &    0.9982  &     101.0  &            &    0.9944    &      14.3  \\

       RC2 &    0.9856  &       1.7  &            &    0.9973  &      59.0  &            &    0.9968   &       7.4  &            &    0.9996  &     164.1  &            &   {\bf 0.9998 }  &      52.1  \\

pr01 -- pr20  &    0.9213  &       9.5  &            &    0.9695  &     224.4  &            &   {\bf 0.9933 }   &     118.3  &            &    0.9928  &     460.5  &            &    0.9895   &   1,680.5  \\

\hline
       $m$ = 4 &            &            &            &            &            &            &                   &            &            &            &            &            &                      &            \\

        C1 &    0.9689  &       2.6  &            &    0.9945  &      49.5  &            &    0.9897   &     133.2  &            &    0.9990  &     261.8  &            &   {\bf 1.0000 }   &      38.6  \\

        C2 &   {\bf 1.0000} &  1.0  &     & {\bf 1.0000}  &      41.8  &     & {\bf 1.0000}    &       0.6  &     &    {\bf 1.0000}  &       0.1  &            & {\bf 1.0000}    &       6.5  \\

        R1 &    0.9670  &       2.6  &            &    0.9929  &      58.4  &            &    0.9880    &      84.7  &            &  {\bf   0.9986}  &     184.3  &            &    0.9956    &      21.0  \\

        R2 &  {\bf 1.0000}  &       0.9  &            &   {\bf  1.0000}  &      39.7  &            &   {\bf  1.0000}    &       0.3  &            &   {\bf  1.0000}  &       0.2  &            &   {\bf  1.0000}    &       7.7  \\

       RC1 &    0.9693  &       2.0  &            &    0.9974  &      68.1  &            &    0.9916    &      36.9  &            &   {\bf  0.9988}  &     152.4  &            &    0.9975   &      17.7  \\

       RC2 &   {\bf  1.0000}  &       1.2  &            &   {\bf  1.0000}  &      40.2  &            &   {\bf  1.0000}    &       0.9  &            &   {\bf  1.0000}  &       0.2  &            &    {\bf 1.0000}    &      15.5  \\

pr01 -- pr20  &    0.9249  &      13.9  &            &    0.9719  &     269.8  &            &    0.9872    &     180.0  &            &   {\bf  0.9988}  &     647.6  &            &    0.9940  &   1,542.4  \\
\hline
   $m$ = opt &            &            &            &            &            &            &                       &            &            &            &            &                    &            &            \\

        C1 &    0.9859  &       3.0  &            &    0.9896  &      77.7  &            &      N/A      &       7.8  &            &   {\bf  1.0000}  &      47.6  &            &  {\bf   1.0000}   &      61.8  \\

        C2 &   {\bf  1.0000}  &       1.1  &            &  {\bf   1.0000}  &      41.9  &            &       N/A        &       0.5  &            &  {\bf   1.0000}  &       0.6  &            &  {\bf   1.0000}   &     113.3  \\

        R1 &    0.9807  &       3.0  &            &    0.9958  &     104.7  &            &      N/A       &      39.5  &            &    0.9993  &     877.7  &            &  {\bf   0.9999 }  &      58.2  \\

        R2 &    0.9938  &       1.7  &            &    0.9984  &      58.3  &            &      N/A        &       5.5  &            &    0.9993  &     173.4  &            &   {\bf  1.0000 }  &     265.2  \\

       RC1 &    0.9794  &       3.8  &            &    0.9965  &      84.6  &            &      N/A         &      39.5  &            &  {\bf   1.0000 } &      57.4  &            &    0.9994   &      54.3  \\

       RC2 &    0.9953  &       1.7  &            &    0.9993  &      41.4  &            &      N/A        &       2.8  &            &    0.9996  &     190.2  &            &   {\bf  1.0000}  &     194.9  \\

pr01 -- pr10  &    0.9768  &      30.4  &            &    0.9896  &     566.0  &            &     N/A          &      51.3  &            &    0.9922  &     326.7  &            &   {\bf  0.9933}  &     420.5  \\
\hline
Overall Average &    0.9751  &       3.2  &            &    0.9933  &      83.3  &            &    0.9914    &      51.6  &            &    0.9957  &     185.4  &            &  {\bf   0.9980}  &     222.2  \\
\hline
    \multicolumn{15}{l}{$m$ = opt implies that these instances belong to category OPT.}

\end{tabular}
}
\end{center}
\end{table*}

The numbers of the best solution values achieved by ILS, SSA, GVNS, I3CH and TS are 85, 191, 138, 247 and 272, respectively (see Appendix B). Although TS produced the largest number of the best solution values and the largest overall average ratio (i.e., 0.9980), we cannot conclude that this algorithm is superior to the rest since it was executed more times and consumed more computation time. We can only say that the results generated by our TS algorithm are comparable to those generated by the best existing approaches for the TOPTW.

\subsection{Analysis of components}
As our TS algorithm consists of three main components, namely local search with tabu moves (LS), ejection pool (EP) algorithm and perturbation (PER) procedure, it is important to investigate the performance of these components. In the experiments, we considered the combinations LS + EP, LS + PER, EP + PER and LS + EP + PER and 50 TOPTW instances generated from pr01 -- pr10. Table \ref{appendix:tab:com} shows the average performance of these four combinations. For each test instance, we calculated a ratio that is equal to the average profit over ten runs divided by the best solution value generated by these four combinations (for the detailed results, see Appendix C). The column ``Avg. Ratio (\%)'' shows the average values of the ratios of the instances grouped by $m$. From this table, we can see that on average, LS + EP + PER performed best while LS + PER generated the worst results. Moreover, EP + PER and LS + EP performed slightly worse than LS + EP + PER. These observations imply that the ejection pool algorithm plays a critical role in improving the solution quality of our proposed approach.

\begin{table*}[htp]
\normalsize{\caption{Average performance of four combinations on the 50 TOPTW instances generated from pr01 - pr10.} \label{appendix:tab:com}}
\begin{center}
\scalebox{0.6}{
\begin{tabular}{cccccccccccc}
\hline
\multirow{2}{*}{$m$} & \multicolumn{ 2}{c}{EP + PER} &            & \multicolumn{ 2}{c}{LS + EP} &            & \multicolumn{ 2}{c}{LS + PER} &            & \multicolumn{ 2}{c}{LS + EP + PER} \\
\cline{2 -3} \cline{5 -6} \cline{8 - 9} \cline{11 - 12}
                 & Avg. Ratio (\%) &  Avg. Time &            & Avg. Ratio (\%) &  Avg. Time &            & Avg. Ratio (\%) &  Avg. Time &            & Avg. Ratio (\%) &  Avg. Time \\
\hline
         1 &    0.9742  &     515.4  &            &    0.9755  &     184.9  &            &    0.9796  &      63.2  &            &    0.9895  &     347.3  \\

         2 &    0.9771  &   1,985.8  &            &    0.9743  &     734.7  &            &    0.9689  &     123.6  &            &    0.9835  &   1,046.7  \\

         3 &    0.9824  &   2,883.0  &            &    0.9799  &     841.5  &            &    0.9688  &     148.1  &            &    0.9832  &   1,122.6  \\

         4 &    0.9860  &   3,956.5  &            &    0.9807  &     892.3  &            &    0.9697  &     179.9  &            &    0.9800  &     978.9  \\

       opt &    0.9977  &   2,169.2  &            &    0.9954  &     196.1  &            &    0.9979  &     401.2  &            &    0.9987  &     420.5  \\
\hline
Overall Average &    0.9835  &   2,302.0  &            &    0.9811  &     569.9  &            &    0.9770  &     183.2  &            &    0.9870  &     783.2  \\
\hline
    \multicolumn{12}{l}{$m$ = opt implies that these instances belong to category OPT.}
\end{tabular}
}
\end{center}
\end{table*}

\subsection{Results for the MPISP instances}
The computational results for the 672 MPISP instances are reported in Tables \ref{tab:mpisp:1} -- \ref{tab:mpisp:4}. The column ``UB'' shows the upper bound of each instance obtained from the constrained knapsack model (see Section \ref{sec:ub}). Each block corresponds to a value of $w$ and includes the maximum workload {\em Max. Workload}, the average workload {\em Ave. Workload} and the average computation time {\em Ave. Time} over the ten executions. Since the MPISP is a new problem, there is no existing algorithm tailored for it. As a consequence, we cannot compare our tabu search algorithm with other approaches. The test instances and computational results reported in this article can serve as benchmarks for future researchers on this problem.

\begin{table*}[htp]
\normalsize{\caption{Computational results for the MPISP instances with $m = 7$.}
\label{tab:mpisp:1}}
\begin{center}
 \scalebox{0.5}{
\begin{tabular}{ccccccccccccccc}
\hline
&                          \multicolumn{ 4}{c}{$w = 1$} &            &                          \multicolumn{ 4}{c}{$w = 3$} &            &                          \multicolumn{ 4}{c}{$w = 5$} \\
\cline{2-5} \cline{7-10} \cline{12-15}
\multicolumn{ 1}{c}{Instance}  &         UB & Max. Workload & Ave. Workload &  Ave. Time &            &         UB & Max. Workload & Ave. Workload &  Ave. Time &            &         UB & Max. Workload & Ave. Workload &  Ave. Time \\
\hline
      c101 &     1,400  &     1,400  &   1,400.0  &       4.9  &            &     1,400  &     1,400  &   1,399.0  &      25.4  &            &     1,350  &     1,240  &   1,239.0  &      26.0  \\

      c102 &     1,400  &     1,400  &   1,400.0  &       7.1  &            &     1,400  &     1,400  &   1,400.0  &       7.7  &            &     1,400  &     1,400  &   1,400.0  &      11.4  \\

      c103 &     1,400  &     1,400  &   1,400.0  &      11.0  &            &     1,400  &     1,400  &   1,400.0  &      12.0  &            &     1,400  &     1,400  &   1,400.0  &      22.5  \\

      c104 &     1,400  &     1,400  &   1,400.0  &      20.5  &            &     1,400  &     1,400  &   1,400.0  &      23.5  &            &     1,400  &     1,400  &   1,400.0  &      33.4  \\

      c105 &     1,400  &     1,400  &   1,400.0  &       5.8  &            &     1,400  &     1,400  &   1,400.0  &       7.0  &            &     1,400  &     1,400  &   1,400.0  &      12.5  \\

      c106 &     1,400  &     1,400  &   1,400.0  &       6.6  &            &     1,400  &     1,400  &   1,400.0  &       6.6  &            &     1,400  &     1,370  &   1,368.0  &      53.3  \\

      c107 &     1,400  &     1,400  &   1,400.0  &       7.6  &            &     1,400  &     1,400  &   1,400.0  &       7.8  &            &     1,400  &     1,400  &   1,400.0  &      12.8  \\

      c108 &     1,400  &     1,400  &   1,400.0  &       8.9  &            &     1,400  &     1,400  &   1,400.0  &      10.7  &            &     1,400  &     1,400  &   1,400.0  &      14.6  \\

      c109 &     1,400  &     1,400  &   1,400.0  &      11.4  &            &     1,400  &     1,400  &   1,400.0  &      12.5  &            &     1,400  &     1,400  &   1,400.0  &      21.1  \\
\hline
      c201 &     1,400  &     1,400  &   1,400.0  &       7.0  &            &     1,400  &     1,400  &   1,400.0  &       7.4  &            &     1,400  &     1,400  &   1,400.0  &       7.8  \\

      c202 &     1,400  &     1,400  &   1,400.0  &      13.3  &            &     1,400  &     1,400  &   1,400.0  &      12.2  &            &     1,400  &     1,400  &   1,400.0  &      14.5  \\

      c203 &     1,400  &     1,400  &   1,400.0  &      19.9  &            &     1,400  &     1,400  &   1,400.0  &      23.1  &            &     1,400  &     1,400  &   1,400.0  &      32.0  \\

      c204 &     1,400  &     1,400  &   1,400.0  &      34.9  &            &     1,400  &     1,400  &   1,400.0  &      35.8  &            &     1,400  &     1,400  &   1,400.0  &      47.6  \\

      c205 &     1,400  &     1,400  &   1,400.0  &       9.1  &            &     1,400  &     1,400  &   1,400.0  &      10.0  &            &     1,400  &     1,400  &   1,400.0  &      10.0  \\

      c206 &     1,400  &     1,400  &   1,400.0  &      13.0  &            &     1,400  &     1,400  &   1,400.0  &      13.2  &            &     1,400  &     1,400  &   1,400.0  &      14.4  \\

      c207 &     1,400  &     1,400  &   1,400.0  &      13.9  &            &     1,400  &     1,400  &   1,400.0  &      15.2  &            &     1,400  &     1,400  &   1,400.0  &      18.2  \\

      c208 &     1,400  &     1,400  &   1,400.0  &      14.8  &            &     1,400  &     1,400  &   1,400.0  &      15.8  &            &     1,400  &     1,400  &   1,400.0  &      17.8  \\
\hline
      r101 &     1,001  &       941  &     934.7  &      27.3  &            &     1,001  &       891  &     889.1  &      36.0  &            &     1,001  &       885  &     879.1  &      51.7  \\

      r102 &     1,222  &     1,146  &   1,140.6  &      47.1  &            &     1,222  &     1,109  &   1,103.4  &     106.7  &            &     1,222  &     1,122  &   1,113.7  &     116.2  \\

      r103 &     1,374  &     1,277  &   1,264.9  &      64.6  &            &     1,374  &     1,243  &   1,233.8  &     102.0  &            &     1,374  &     1,231  &   1,223.5  &      77.1  \\

      r104 &     1,400  &     1,335  &   1,323.1  &      75.2  &            &     1,400  &     1,307  &   1,288.4  &      84.2  &            &     1,400  &     1,290  &   1,277.2  &      88.7  \\

      r105 &     1,400  &     1,128  &   1,115.9  &      46.3  &            &     1,400  &     1,095  &   1,086.6  &      53.5  &            &     1,400  &     1,071  &   1,062.8  &      66.5  \\

      r106 &     1,400  &     1,247  &   1,236.2  &      43.7  &            &     1,400  &     1,230  &   1,218.8  &      68.9  &            &     1,400  &     1,210  &   1,198.4  &     109.5  \\

      r107 &     1,400  &     1,302  &   1,290.1  &      53.6  &            &     1,400  &     1,284  &   1,268.0  &      86.1  &            &     1,400  &     1,262  &   1,250.9  &     103.6  \\

      r108 &     1,400  &     1,342  &   1,330.2  &      51.7  &            &     1,400  &     1,331  &   1,317.6  &      72.8  &            &     1,400  &     1,311  &   1,301.6  &     131.1  \\

      r109 &     1,400  &     1,235  &   1,222.9  &      43.4  &            &     1,400  &     1,218  &   1,204.1  &      60.3  &            &     1,400  &     1,190  &   1,175.7  &      83.9  \\

      r110 &     1,400  &     1,295  &   1,279.8  &      54.5  &            &     1,400  &     1,253  &   1,241.2  &      56.6  &            &     1,400  &     1,247  &   1,235.4  &      74.2  \\

      r111 &     1,400  &     1,295  &   1,285.5  &      46.7  &            &     1,400  &     1,274  &   1,263.5  &      79.3  &            &     1,400  &     1,259  &   1,246.4  &      90.4  \\

      r112 &     1,400  &     1,343  &   1,329.1  &      59.8  &            &     1,400  &     1,318  &   1,306.9  &      90.6  &            &     1,400  &     1,319  &   1,300.7  &      73.4  \\
\hline
      r201 &     1,400  &     1,400  &   1,400.0  &       5.8  &            &     1,400  &     1,400  &   1,400.0  &       6.1  &            &     1,400  &     1,400  &   1,400.0  &       7.4  \\

      r202 &     1,400  &     1,400  &   1,400.0  &       6.4  &            &     1,400  &     1,400  &   1,400.0  &       6.5  &            &     1,400  &     1,400  &   1,400.0  &       7.8  \\

      r203 &     1,400  &     1,400  &   1,400.0  &       8.4  &            &     1,400  &     1,400  &   1,400.0  &       7.3  &            &     1,400  &     1,400  &   1,400.0  &       9.1  \\

      r204 &     1,400  &     1,400  &   1,400.0  &       9.5  &            &     1,400  &     1,400  &   1,400.0  &      10.4  &            &     1,400  &     1,400  &   1,400.0  &      11.5  \\

      r205 &     1,400  &     1,400  &   1,400.0  &       7.6  &            &     1,400  &     1,400  &   1,400.0  &       9.6  &            &     1,400  &     1,400  &   1,400.0  &       9.0  \\

      r206 &     1,400  &     1,400  &   1,400.0  &       8.5  &            &     1,400  &     1,400  &   1,400.0  &       9.8  &            &     1,400  &     1,400  &   1,400.0  &      11.0  \\

      r207 &     1,400  &     1,400  &   1,400.0  &       9.0  &            &     1,400  &     1,400  &   1,400.0  &      11.3  &            &     1,400  &     1,400  &   1,400.0  &      12.0  \\

      r208 &     1,400  &     1,400  &   1,400.0  &      12.0  &            &     1,400  &     1,400  &   1,400.0  &      14.0  &            &     1,400  &     1,400  &   1,400.0  &      15.4  \\

      r209 &     1,400  &     1,400  &   1,400.0  &       8.8  &            &     1,400  &     1,400  &   1,400.0  &      10.7  &            &     1,400  &     1,400  &   1,400.0  &      11.5  \\

      r210 &     1,400  &     1,400  &   1,400.0  &       7.5  &            &     1,400  &     1,400  &   1,400.0  &       8.9  &            &     1,400  &     1,400  &   1,400.0  &       9.5  \\

      r211 &     1,400  &     1,400  &   1,400.0  &      12.9  &            &     1,400  &     1,400  &   1,400.0  &      16.2  &            &     1,400  &     1,400  &   1,400.0  &      18.7  \\
\hline
     rc101 &     1,400  &     1,228  &   1,211.1  &      33.0  &            &     1,400  &     1,174  &   1,166.5  &      40.6  &            &     1,400  &     1,161  &   1,149.7  &      66.2  \\

     rc102 &     1,400  &     1,359  &   1,350.0  &      50.7  &            &     1,400  &     1,339  &   1,326.6  &      56.5  &            &     1,400  &     1,331  &   1,312.6  &      69.4  \\

     rc103 &     1,400  &     1,399  &   1,383.1  &      61.0  &            &     1,400  &     1,393  &   1,377.0  &      75.5  &            &     1,400  &     1,368  &   1,359.0  &      86.7  \\

     rc104 &     1,400  &     1,400  &   1,396.6  &      61.1  &            &     1,400  &     1,400  &   1,395.6  &      79.4  &            &     1,400  &     1,398  &   1,387.8  &     106.8  \\

     rc105 &     1,400  &     1,331  &   1,321.2  &      43.5  &            &     1,400  &     1,309  &   1,296.3  &      49.8  &            &     1,400  &     1,282  &   1,275.2  &      57.3  \\

     rc106 &     1,400  &     1,369  &   1,344.5  &      60.2  &            &     1,400  &     1,329  &   1,315.2  &      76.7  &            &     1,400  &     1,310  &   1,296.1  &      89.5  \\

     rc107 &     1,400  &     1,393  &   1,379.2  &      61.5  &            &     1,400  &     1,361  &   1,348.4  &      91.6  &            &     1,400  &     1,369  &   1,347.9  &     198.6  \\

     rc108 &     1,400  &     1,399  &   1,390.0  &      38.1  &            &     1,400  &     1,388  &   1,375.4  &      95.4  &            &     1,400  &     1,370  &   1,359.3  &     165.1  \\
\hline
     rc201 &     1,400  &     1,400  &   1,400.0  &       8.6  &            &     1,400  &     1,400  &   1,400.0  &      10.5  &            &     1,400  &     1,400  &   1,400.0  &      12.1  \\

     rc202 &     1,400  &     1,400  &   1,400.0  &       9.1  &            &     1,400  &     1,400  &   1,400.0  &      11.0  &            &     1,400  &     1,400  &   1,400.0  &      11.7  \\

     rc203 &     1,400  &     1,400  &   1,400.0  &      12.3  &            &     1,400  &     1,400  &   1,400.0  &      13.9  &            &     1,400  &     1,400  &   1,400.0  &      16.2  \\

     rc204 &     1,400  &     1,400  &   1,400.0  &      18.2  &            &     1,400  &     1,400  &   1,400.0  &      28.7  &            &     1,400  &     1,400  &   1,400.0  &      27.9  \\

     rc205 &     1,400  &     1,400  &   1,400.0  &      10.4  &            &     1,400  &     1,400  &   1,400.0  &      11.0  &            &     1,400  &     1,400  &   1,400.0  &      10.8  \\

     rc206 &     1,400  &     1,400  &   1,400.0  &      13.2  &            &     1,400  &     1,400  &   1,400.0  &      16.3  &            &     1,400  &     1,400  &   1,400.0  &      18.1  \\

     rc207 &     1,400  &     1,400  &   1,400.0  &      16.0  &            &     1,400  &     1,400  &   1,400.0  &      18.1  &            &     1,400  &     1,400  &   1,400.0  &      19.8  \\

     rc208 &     1,400  &     1,400  &   1,400.0  &      22.9  &            &     1,400  &     1,400  &   1,400.0  &      28.1  &            &     1,400  &     1,400  &   1,400.0  &      29.2  \\
\hline
\end{tabular}
}
\end{center}
\end{table*}

\begin{table*}[htp]
\normalsize{\caption{Computational results for the MPISP instances with $m = 9$.}
\label{tab:mpisp:2}}
\begin{center}
 \scalebox{0.5}{
\begin{tabular}{ccccccccccccccc}
\hline
&                          \multicolumn{ 4}{c}{$w = 1$} &            &                          \multicolumn{ 4}{c}{$w = 3$} &            &                          \multicolumn{ 4}{c}{$w = 5$} \\
\cline{2-5} \cline{7-10} \cline{12-15}
\multicolumn{ 1}{c}{Instance}  &         UB & Max. Workload & Ave. Workload &  Ave. Time &            &         UB & Max. Workload & Ave. Workload &  Ave. Time &            &         UB & Max. Workload & Ave. Workload &  Ave. Time \\
\hline
      c101 &     1,730  &     1,710  &   1,707.0  &      22.5  &            &     1,670  &     1,630  &   1,621.0  &      18.3  &            &     1,480  &     1,380  &   1,380.0  &       6.2  \\

      c102 &     1,800  &     1,800  &   1,799.0  &      24.6  &            &     1,760  &     1,750  &   1,748.0  &      30.3  &            &     1,640  &     1,620  &   1,620.0  &       9.8  \\

      c103 &     1,800  &     1,800  &   1,800.0  &       9.9  &            &     1,780  &     1,770  &   1,770.0  &      11.5  &            &     1,680  &     1,680  &   1,680.0  &       8.7  \\

      c104 &     1,800  &     1,800  &   1,800.0  &       5.5  &            &     1,800  &     1,800  &   1,800.0  &      16.2  &            &     1,680  &     1,680  &   1,680.0  &      13.9  \\

      c105 &     1,800  &     1,730  &   1,728.0  &      31.6  &            &     1,800  &     1,680  &   1,677.0  &      22.5  &            &     1,710  &     1,590  &   1,590.0  &      11.1  \\

      c106 &     1,800  &     1,740  &   1,738.0  &      32.6  &            &     1,780  &     1,690  &   1,690.0  &      23.3  &            &     1,650  &     1,530  &   1,530.0  &      10.1  \\

      c107 &     1,800  &     1,750  &   1,748.0  &      52.5  &            &     1,800  &     1,720  &   1,718.0  &      46.7  &            &     1,710  &     1,600  &   1,600.0  &      12.0  \\

      c108 &     1,800  &     1,780  &   1,770.0  &      34.0  &            &     1,800  &     1,740  &   1,739.0  &      28.1  &            &     1,710  &     1,610  &   1,610.0  &      14.6  \\

      c109 &     1,800  &     1,800  &   1,800.0  &      22.8  &            &     1,800  &     1,800  &   1,796.0  &      39.2  &            &     1,710  &     1,650  &   1,650.0  &      15.5  \\
\hline
      c201 &     1,800  &     1,800  &   1,800.0  &       4.3  &            &     1,800  &     1,800  &   1,800.0  &       4.5  &            &     1,800  &     1,800  &   1,800.0  &       4.7  \\

      c202 &     1,800  &     1,800  &   1,800.0  &       5.0  &            &     1,800  &     1,800  &   1,800.0  &       5.1  &            &     1,800  &     1,800  &   1,800.0  &       4.8  \\

      c203 &     1,800  &     1,800  &   1,800.0  &       5.6  &            &     1,800  &     1,800  &   1,800.0  &       5.8  &            &     1,800  &     1,800  &   1,800.0  &       5.9  \\

      c204 &     1,800  &     1,800  &   1,800.0  &       6.5  &            &     1,800  &     1,800  &   1,800.0  &       6.7  &            &     1,800  &     1,800  &   1,800.0  &       6.9  \\

      c205 &     1,800  &     1,800  &   1,800.0  &       4.6  &            &     1,800  &     1,800  &   1,800.0  &       5.0  &            &     1,800  &     1,800  &   1,800.0  &       4.6  \\

      c206 &     1,800  &     1,800  &   1,800.0  &       4.9  &            &     1,800  &     1,800  &   1,800.0  &       5.0  &            &     1,800  &     1,800  &   1,800.0  &       5.1  \\

      c207 &     1,800  &     1,800  &   1,800.0  &       5.5  &            &     1,800  &     1,800  &   1,800.0  &       5.3  &            &     1,800  &     1,800  &   1,800.0  &       5.4  \\

      c208 &     1,800  &     1,800  &   1,800.0  &       5.7  &            &     1,800  &     1,800  &   1,800.0  &       5.3  &            &     1,800  &     1,800  &   1,800.0  &       5.5  \\
\hline
      r101 &     1,171  &     1,109  &   1,098.3  &      37.4  &            &     1,171  &     1,045  &   1,042.6  &      86.3  &            &     1,171  &     1,053  &   1,048.8  &      62.7  \\

      r102 &     1,321  &     1,292  &   1,287.7  &      81.1  &            &     1,321  &     1,247  &   1,239.3  &     111.9  &            &     1,321  &     1,264  &   1,259.9  &      92.4  \\

      r103 &     1,413  &     1,395  &   1,388.2  &      69.6  &            &     1,413  &     1,370  &   1,360.4  &      99.4  &            &     1,413  &     1,364  &   1,359.6  &     103.6  \\

      r104 &     1,458  &     1,449  &   1,445.4  &      65.0  &            &     1,458  &     1,414  &   1,408.8  &      75.4  &            &     1,458  &     1,426  &   1,418.9  &      89.3  \\

      r105 &     1,458  &     1,290  &   1,273.8  &      42.0  &            &     1,458  &     1,270  &   1,245.8  &      57.4  &            &     1,458  &     1,228  &   1,218.3  &      95.8  \\

      r106 &     1,458  &     1,386  &   1,371.5  &      43.8  &            &     1,458  &     1,359  &   1,347.9  &      65.6  &            &     1,458  &     1,350  &   1,340.0  &     112.0  \\

      r107 &     1,458  &     1,429  &   1,419.8  &      46.2  &            &     1,458  &     1,419  &   1,409.0  &      84.3  &            &     1,458  &     1,397  &   1,390.8  &      99.3  \\

      r108 &     1,458  &     1,458  &   1,457.5  &      71.5  &            &     1,458  &     1,455  &   1,446.5  &      62.4  &            &     1,458  &     1,445  &   1,436.4  &     119.7  \\

      r109 &     1,458  &     1,378  &   1,371.1  &      33.0  &            &     1,458  &     1,367  &   1,358.0  &      45.8  &            &     1,458  &     1,340  &   1,330.7  &      97.1  \\

      r110 &     1,458  &     1,424  &   1,416.0  &      46.3  &            &     1,458  &     1,407  &   1,395.0  &      90.9  &            &     1,458  &     1,388  &   1,373.4  &      81.5  \\

      r111 &     1,458  &     1,433  &   1,426.3  &      69.2  &            &     1,458  &     1,411  &   1,399.7  &      51.9  &            &     1,458  &     1,401  &   1,394.1  &      61.4  \\

      r112 &     1,458  &     1,458  &   1,455.4  &      65.4  &            &     1,458  &     1,452  &   1,446.1  &      96.3  &            &     1,458  &     1,440  &   1,430.3  &      86.5  \\
\hline
      r201 &     1,458  &     1,458  &   1,458.0  &       6.1  &            &     1,458  &     1,458  &   1,458.0  &       5.9  &            &     1,458  &     1,458  &   1,458.0  &       6.5  \\

      r202 &     1,458  &     1,458  &   1,458.0  &       6.5  &            &     1,458  &     1,458  &   1,458.0  &       6.6  &            &     1,458  &     1,458  &   1,458.0  &       7.0  \\

      r203 &     1,458  &     1,458  &   1,458.0  &       7.4  &            &     1,458  &     1,458  &   1,458.0  &       7.4  &            &     1,458  &     1,458  &   1,458.0  &       7.8  \\

      r204 &     1,458  &     1,458  &   1,458.0  &       8.7  &            &     1,458  &     1,458  &   1,458.0  &       8.8  &            &     1,458  &     1,458  &   1,458.0  &       9.3  \\

      r205 &     1,458  &     1,458  &   1,458.0  &       7.1  &            &     1,458  &     1,458  &   1,458.0  &       7.4  &            &     1,458  &     1,458  &   1,458.0  &       7.9  \\

      r206 &     1,458  &     1,458  &   1,458.0  &       7.4  &            &     1,458  &     1,458  &   1,458.0  &       7.8  &            &     1,458  &     1,458  &   1,458.0  &       8.2  \\

      r207 &     1,458  &     1,458  &   1,458.0  &       9.1  &            &     1,458  &     1,458  &   1,458.0  &       8.9  &            &     1,458  &     1,458  &   1,458.0  &       9.6  \\

      r208 &     1,458  &     1,458  &   1,458.0  &      10.9  &            &     1,458  &     1,458  &   1,458.0  &      11.0  &            &     1,458  &     1,458  &   1,458.0  &      11.7  \\

      r209 &     1,458  &     1,458  &   1,458.0  &       7.3  &            &     1,458  &     1,458  &   1,458.0  &       7.9  &            &     1,458  &     1,458  &   1,458.0  &       8.1  \\

      r210 &     1,458  &     1,458  &   1,458.0  &       6.9  &            &     1,458  &     1,458  &   1,458.0  &       7.9  &            &     1,458  &     1,458  &   1,458.0  &       8.5  \\

      r211 &     1,458  &     1,458  &   1,458.0  &      10.0  &            &     1,458  &     1,458  &   1,458.0  &       9.7  &            &     1,458  &     1,458  &   1,458.0  &      10.8  \\
\hline
     rc101 &     1,724  &     1,456  &   1,435.9  &      51.9  &            &     1,724  &     1,391  &   1,380.2  &      78.3  &            &     1,724  &     1,371  &   1,362.4  &      91.6  \\

     rc102 &     1,724  &     1,585  &   1,575.2  &      41.4  &            &     1,724  &     1,567  &   1,556.4  &      75.9  &            &     1,724  &     1,535  &   1,523.1  &      86.6  \\

     rc103 &     1,724  &     1,672  &   1,655.3  &      52.9  &            &     1,724  &     1,651  &   1,630.6  &      90.6  &            &     1,724  &     1,635  &   1,612.0  &      79.6  \\

     rc104 &     1,724  &     1,702  &   1,689.1  &      47.7  &            &     1,724  &     1,673  &   1,663.0  &      58.4  &            &     1,724  &     1,668  &   1,647.7  &     116.9  \\

     rc105 &     1,696  &     1,542  &   1,533.8  &      45.3  &            &     1,696  &     1,527  &   1,515.3  &      54.2  &            &     1,696  &     1,486  &   1,477.9  &      68.5  \\

     rc106 &     1,724  &     1,606  &   1,586.7  &      40.4  &            &     1,724  &     1,577  &   1,563.2  &      85.8  &            &     1,724  &     1,552  &   1,539.3  &     104.6  \\

     rc107 &     1,724  &     1,651  &   1,633.5  &      59.0  &            &     1,724  &     1,637  &   1,614.2  &      87.4  &            &     1,724  &     1,613  &   1,596.5  &      92.2  \\

     rc108 &     1,724  &     1,687  &   1,666.1  &      41.2  &            &     1,724  &     1,661  &   1,642.4  &      52.4  &            &     1,724  &     1,658  &   1,643.4  &     114.3  \\
\hline
     rc201 &     1,724  &     1,724  &   1,724.0  &       5.1  &            &     1,724  &     1,724  &   1,724.0  &       5.1  &            &     1,724  &     1,724  &   1,724.0  &       5.5  \\

     rc202 &     1,724  &     1,724  &   1,724.0  &       5.6  &            &     1,724  &     1,724  &   1,724.0  &       5.7  &            &     1,724  &     1,724  &   1,724.0  &       4.5  \\

     rc203 &     1,724  &     1,724  &   1,724.0  &       5.9  &            &     1,724  &     1,724  &   1,724.0  &       5.7  &            &     1,724  &     1,724  &   1,724.0  &       5.6  \\

     rc204 &     1,724  &     1,724  &   1,724.0  &       7.4  &            &     1,724  &     1,724  &   1,724.0  &       7.1  &            &     1,724  &     1,724  &   1,724.0  &       7.5  \\

     rc205 &     1,724  &     1,724  &   1,724.0  &       5.4  &            &     1,724  &     1,724  &   1,724.0  &       5.0  &            &     1,724  &     1,724  &   1,724.0  &       4.1  \\

     rc206 &     1,724  &     1,724  &   1,724.0  &       5.9  &            &     1,724  &     1,724  &   1,724.0  &       5.6  &            &     1,724  &     1,724  &   1,724.0  &       5.2  \\

     rc207 &     1,724  &     1,724  &   1,724.0  &       6.4  &            &     1,724  &     1,724  &   1,724.0  &       5.7  &            &     1,724  &     1,724  &   1,724.0  &       5.6  \\

     rc208 &     1,724  &     1,724  &   1,724.0  &       8.1  &            &     1,724  &     1,724  &   1,724.0  &       7.6  &            &     1,724  &     1,724  &   1,724.0  &       6.3  \\
\hline
\end{tabular}
}
\end{center}
\end{table*}

\begin{table*}[htp]
\normalsize{\caption{Computational results for the MPISP instances with $m = 11$.}
\label{tab:mpisp:3}}
\begin{center}
 \scalebox{0.5}{
\begin{tabular}{ccccccccccccccc}
\hline
&                          \multicolumn{ 4}{c}{$w = 1$} &            &                          \multicolumn{ 4}{c}{$w = 3$} &            &                          \multicolumn{ 4}{c}{$w = 5$} \\
\cline{2-5} \cline{7-10} \cline{12-15}
\multicolumn{ 1}{c}{Instance}  &         UB & Max. Workload & Ave. Workload &  Ave. Time &            &         UB & Max. Workload & Ave. Workload &  Ave. Time &            &         UB & Max. Workload & Ave. Workload &  Ave. Time \\
\hline
      c101 &     1,810  &     1,810  &   1,810.0  &       4.0  &            &     1,740  &     1,740  &   1,740.0  &      18.2  &            &     1,540  &     1,480  &   1,480.0  &       4.2  \\

      c102 &     1,810  &     1,810  &   1,810.0  &       4.9  &            &     1,760  &     1,760  &   1,760.0  &       4.5  &            &     1,670  &     1,670  &   1,670.0  &      11.3  \\

      c103 &     1,810  &     1,810  &   1,810.0  &       5.4  &            &     1,780  &     1,780  &   1,780.0  &       5.2  &            &     1,730  &     1,730  &   1,730.0  &       6.3  \\

      c104 &     1,810  &     1,810  &   1,810.0  &       6.9  &            &     1,800  &     1,800  &   1,800.0  &       6.7  &            &     1,750  &     1,750  &   1,750.0  &       8.2  \\

      c105 &     1,810  &     1,810  &   1,810.0  &       4.0  &            &     1,810  &     1,810  &   1,810.0  &       8.4  &            &     1,810  &     1,730  &   1,730.0  &       7.2  \\

      c106 &     1,810  &     1,810  &   1,810.0  &       4.7  &            &     1,780  &     1,780  &   1,780.0  &       4.7  &            &     1,710  &     1,670  &   1,670.0  &       7.0  \\

      c107 &     1,810  &     1,810  &   1,810.0  &       4.8  &            &     1,810  &     1,810  &   1,810.0  &       4.6  &            &     1,810  &     1,760  &   1,760.0  &       8.7  \\

      c108 &     1,810  &     1,810  &   1,810.0  &       5.1  &            &     1,810  &     1,810  &   1,810.0  &       4.8  &            &     1,810  &     1,770  &   1,770.0  &      14.7  \\

      c109 &     1,810  &     1,810  &   1,810.0  &       6.7  &            &     1,810  &     1,810  &   1,810.0  &       5.9  &            &     1,810  &     1,810  &   1,810.0  &       7.7  \\
\hline
      c201 &     1,810  &     1,810  &   1,810.0  &       4.8  &            &     1,810  &     1,810  &   1,810.0  &       5.0  &            &     1,810  &     1,810  &   1,810.0  &       5.0  \\

      c202 &     1,810  &     1,810  &   1,810.0  &       5.4  &            &     1,810  &     1,810  &   1,810.0  &       5.4  &            &     1,810  &     1,810  &   1,810.0  &       5.6  \\

      c203 &     1,810  &     1,810  &   1,810.0  &       6.5  &            &     1,810  &     1,810  &   1,810.0  &       6.5  &            &     1,810  &     1,810  &   1,810.0  &       6.3  \\

      c204 &     1,810  &     1,810  &   1,810.0  &       8.2  &            &     1,810  &     1,810  &   1,810.0  &       7.8  &            &     1,810  &     1,810  &   1,810.0  &       8.3  \\

      c205 &     1,810  &     1,810  &   1,810.0  &       5.2  &            &     1,810  &     1,810  &   1,810.0  &       5.2  &            &     1,810  &     1,810  &   1,810.0  &       5.3  \\

      c206 &     1,810  &     1,810  &   1,810.0  &       5.3  &            &     1,810  &     1,810  &   1,810.0  &       5.6  &            &     1,810  &     1,810  &   1,810.0  &       5.6  \\

      c207 &     1,810  &     1,810  &   1,810.0  &       5.9  &            &     1,810  &     1,810  &   1,810.0  &       6.3  &            &     1,810  &     1,810  &   1,810.0  &       5.9  \\

      c208 &     1,810  &     1,810  &   1,810.0  &       6.0  &            &     1,810  &     1,810  &   1,810.0  &       6.0  &            &     1,810  &     1,810  &   1,810.0  &       6.5  \\
\hline
      r101 &     1,307  &     1,243  &   1,225.5  &      28.2  &            &     1,307  &     1,170  &   1,158.2  &      58.8  &            &     1,307  &     1,191  &   1,181.3  &      51.3  \\

      r102 &     1,388  &     1,367  &   1,358.3  &      38.9  &            &     1,388  &     1,328  &   1,321.6  &      62.0  &            &     1,388  &     1,338  &   1,333.5  &      80.2  \\

      r103 &     1,441  &     1,441  &   1,433.6  &      22.2  &            &     1,441  &     1,423  &   1,420.0  &      46.9  &            &     1,441  &     1,425  &   1,421.9  &      77.3  \\

      r104 &     1,458  &     1,458  &   1,458.0  &       5.3  &            &     1,458  &     1,458  &   1,452.8  &      29.1  &            &     1,458  &     1,458  &   1,458.0  &      27.4  \\

      r105 &     1,458  &     1,391  &   1,379.0  &      25.0  &            &     1,458  &     1,378  &   1,363.4  &      57.8  &            &     1,458  &     1,351  &   1,339.2  &      61.7  \\

      r106 &     1,458  &     1,453  &   1,445.5  &      53.4  &            &     1,458  &     1,435  &   1,426.7  &      54.9  &            &     1,458  &     1,432  &   1,422.0  &      83.7  \\

      r107 &     1,458  &     1,458  &   1,458.0  &       5.4  &            &     1,458  &     1,458  &   1,458.0  &      14.0  &            &     1,458  &     1,458  &   1,458.0  &      37.0  \\

      r108 &     1,458  &     1,458  &   1,458.0  &       5.2  &            &     1,458  &     1,458  &   1,458.0  &       6.4  &            &     1,458  &     1,458  &   1,458.0  &       6.5  \\

      r109 &     1,458  &     1,458  &   1,455.7  &      39.5  &            &     1,458  &     1,456  &   1,445.4  &      36.3  &            &     1,458  &     1,441  &   1,432.4  &      71.8  \\

      r110 &     1,458  &     1,458  &   1,458.0  &       7.3  &            &     1,458  &     1,458  &   1,458.0  &      20.8  &            &     1,458  &     1,458  &   1,457.4  &      54.0  \\

      r111 &     1,458  &     1,458  &   1,458.0  &       6.4  &            &     1,458  &     1,458  &   1,456.8  &      14.7  &            &     1,458  &     1,458  &   1,457.2  &      39.8  \\

      r112 &     1,458  &     1,458  &   1,458.0  &       5.1  &            &     1,458  &     1,458  &   1,458.0  &       5.7  &            &     1,458  &     1,458  &   1,458.0  &       7.1  \\
\hline
      r201 &     1,458  &     1,458  &   1,458.0  &       6.2  &            &     1,458  &     1,458  &   1,458.0  &       6.2  &            &     1,458  &     1,458  &   1,458.0  &       6.7  \\

      r202 &     1,458  &     1,458  &   1,458.0  &       6.5  &            &     1,458  &     1,458  &   1,458.0  &       6.9  &            &     1,458  &     1,458  &   1,458.0  &       7.1  \\

      r203 &     1,458  &     1,458  &   1,458.0  &       7.4  &            &     1,458  &     1,458  &   1,458.0  &       7.4  &            &     1,458  &     1,458  &   1,458.0  &       8.2  \\

      r204 &     1,458  &     1,458  &   1,458.0  &       8.6  &            &     1,458  &     1,458  &   1,458.0  &       9.4  &            &     1,458  &     1,458  &   1,458.0  &      10.1  \\

      r205 &     1,458  &     1,458  &   1,458.0  &       7.2  &            &     1,458  &     1,458  &   1,458.0  &       7.7  &            &     1,458  &     1,458  &   1,458.0  &       8.0  \\

      r206 &     1,458  &     1,458  &   1,458.0  &       7.4  &            &     1,458  &     1,458  &   1,458.0  &       8.1  &            &     1,458  &     1,458  &   1,458.0  &       8.7  \\

      r207 &     1,458  &     1,458  &   1,458.0  &       9.8  &            &     1,458  &     1,458  &   1,458.0  &       9.5  &            &     1,458  &     1,458  &   1,458.0  &      10.0  \\

      r208 &     1,458  &     1,458  &   1,458.0  &      10.8  &            &     1,458  &     1,458  &   1,458.0  &      11.3  &            &     1,458  &     1,458  &   1,458.0  &      11.9  \\

      r209 &     1,458  &     1,458  &   1,458.0  &       8.3  &            &     1,458  &     1,458  &   1,458.0  &       8.6  &            &     1,458  &     1,458  &   1,458.0  &       8.7  \\

      r210 &     1,458  &     1,458  &   1,458.0  &       8.1  &            &     1,458  &     1,458  &   1,458.0  &       8.0  &            &     1,458  &     1,458  &   1,458.0  &       8.7  \\

      r211 &     1,458  &     1,458  &   1,458.0  &      10.3  &            &     1,458  &     1,458  &   1,458.0  &      10.1  &            &     1,458  &     1,458  &   1,458.0  &      10.7  \\
\hline
     rc101 &     1,724  &     1,621  &   1,609.4  &      41.3  &            &     1,724  &     1,563  &   1,548.6  &      46.2  &            &     1,724  &     1,538  &   1,518.5  &      88.9  \\

     rc102 &     1,724  &     1,690  &   1,677.0  &      33.5  &            &     1,724  &     1,674  &   1,653.7  &      34.3  &            &     1,724  &     1,656  &   1,644.6  &      39.7  \\

     rc103 &     1,724  &     1,724  &   1,723.0  &      21.3  &            &     1,724  &     1,724  &   1,717.2  &      50.5  &            &     1,724  &     1,714  &   1,707.2  &      25.8  \\

     rc104 &     1,724  &     1,724  &   1,724.0  &       7.3  &            &     1,724  &     1,724  &   1,724.0  &       7.0  &            &     1,724  &     1,724  &   1,724.0  &      17.5  \\

     rc105 &     1,719  &     1,681  &   1,673.1  &      48.2  &            &     1,719  &     1,661  &   1,642.6  &      49.2  &            &     1,719  &     1,645  &   1,630.2  &      80.3  \\

     rc106 &     1,724  &     1,724  &   1,714.5  &      55.2  &            &     1,724  &     1,707  &   1,691.6  &      60.1  &            &     1,724  &     1,687  &   1,671.8  &      59.2  \\

     rc107 &     1,724  &     1,724  &   1,724.0  &      14.3  &            &     1,724  &     1,724  &   1,722.7  &      37.1  &            &     1,724  &     1,724  &   1,715.7  &      51.6  \\

     rc108 &     1,724  &     1,724  &   1,724.0  &       4.6  &            &     1,724  &     1,724  &   1,724.0  &       9.3  &            &     1,724  &     1,724  &   1,723.4  &      22.3  \\
\hline
     rc201 &     1,724  &     1,724  &   1,724.0  &       5.5  &            &     1,724  &     1,724  &   1,724.0  &       5.5  &            &     1,724  &     1,724  &   1,724.0  &       5.3  \\

     rc202 &     1,724  &     1,724  &   1,724.0  &       6.2  &            &     1,724  &     1,724  &   1,724.0  &       6.0  &            &     1,724  &     1,724  &   1,724.0  &       5.0  \\

     rc203 &     1,724  &     1,724  &   1,724.0  &       6.1  &            &     1,724  &     1,724  &   1,724.0  &       6.0  &            &     1,724  &     1,724  &   1,724.0  &       6.5  \\

     rc204 &     1,724  &     1,724  &   1,724.0  &       8.4  &            &     1,724  &     1,724  &   1,724.0  &       8.0  &            &     1,724  &     1,724  &   1,724.0  &       8.1  \\

     rc205 &     1,724  &     1,724  &   1,724.0  &       5.4  &            &     1,724  &     1,724  &   1,724.0  &       5.7  &            &     1,724  &     1,724  &   1,724.0  &       4.9  \\

     rc206 &     1,724  &     1,724  &   1,724.0  &       6.0  &            &     1,724  &     1,724  &   1,724.0  &       5.9  &            &     1,724  &     1,724  &   1,724.0  &       5.4  \\

     rc207 &     1,724  &     1,724  &   1,724.0  &       7.1  &            &     1,724  &     1,724  &   1,724.0  &       5.9  &            &     1,724  &     1,724  &   1,724.0  &       6.2  \\

     rc208 &     1,724  &     1,724  &   1,724.0  &       8.4  &            &     1,724  &     1,724  &   1,724.0  &       7.8  &            &     1,724  &     1,724  &   1,724.0  &       6.9  \\
\hline
\end{tabular}
}
\end{center}
\end{table*}

\begin{table*}[htp]
\normalsize{\caption{Computational results for the MPISP instances with $m = 13$.}
\label{tab:mpisp:4}}
\begin{center}
 \scalebox{0.5}{
\begin{tabular}{ccccccccccccccc}
\hline
&                          \multicolumn{ 4}{c}{$w = 1$} &            &                          \multicolumn{ 4}{c}{$w = 3$} &            &                          \multicolumn{ 4}{c}{w=5} \\
\cline{2-5} \cline{7-10} \cline{12-15}
\multicolumn{ 1}{c}{Instance}  &         UB & Max. Workload & Ave. Workload &  Ave. Time &            &         UB & Max. Workload & Ave. Workload &  Ave. Time &            &         UB & Max. Workload & Ave. Workload &  Ave. Time \\
\hline
      c101 &     1,810  &     1,810  &   1,810.0  &       4.2  &            &     1,740  &     1,740  &   1,740.0  &       3.6  &            &     1,540  &     1,530  &   1,530.0  &       3.3  \\

      c102 &     1,810  &     1,810  &   1,810.0  &       5.0  &            &     1,760  &     1,760  &   1,760.0  &       4.2  &            &     1,670  &     1,670  &   1,670.0  &       4.2  \\

      c103 &     1,810  &     1,810  &   1,810.0  &       5.7  &            &     1,780  &     1,780  &   1,780.0  &       5.1  &            &     1,730  &     1,730  &   1,730.0  &       6.4  \\

      c104 &     1,810  &     1,810  &   1,810.0  &       7.1  &            &     1,800  &     1,800  &   1,800.0  &       7.1  &            &     1,750  &     1,750  &   1,750.0  &       9.4  \\

      c105 &     1,810  &     1,810  &   1,810.0  &       4.7  &            &     1,810  &     1,810  &   1,810.0  &       4.7  &            &     1,810  &     1,810  &   1,810.0  &       4.6  \\

      c106 &     1,810  &     1,810  &   1,810.0  &       4.6  &            &     1,780  &     1,780  &   1,780.0  &       4.4  &            &     1,710  &     1,710  &   1,710.0  &       4.0  \\

      c107 &     1,810  &     1,810  &   1,810.0  &       5.2  &            &     1,810  &     1,810  &   1,810.0  &       4.9  &            &     1,810  &     1,810  &   1,810.0  &       4.4  \\

      c108 &     1,810  &     1,810  &   1,810.0  &       5.8  &            &     1,810  &     1,810  &   1,810.0  &       5.4  &            &     1,810  &     1,810  &   1,810.0  &       5.3  \\

      c109 &     1,810  &     1,810  &   1,810.0  &       7.2  &            &     1,810  &     1,810  &   1,810.0  &       6.5  &            &     1,810  &     1,810  &   1,810.0  &       6.1  \\
\hline
      c201 &     1,810  &     1,810  &   1,810.0  &       5.1  &            &     1,810  &     1,810  &   1,810.0  &       5.1  &            &     1,810  &     1,810  &   1,810.0  &       5.1  \\

      c202 &     1,810  &     1,810  &   1,810.0  &       5.4  &            &     1,810  &     1,810  &   1,810.0  &       5.8  &            &     1,810  &     1,810  &   1,810.0  &       5.9  \\

      c203 &     1,810  &     1,810  &   1,810.0  &       6.8  &            &     1,810  &     1,810  &   1,810.0  &       6.6  &            &     1,810  &     1,810  &   1,810.0  &       6.7  \\

      c204 &     1,810  &     1,810  &   1,810.0  &       8.8  &            &     1,810  &     1,810  &   1,810.0  &       7.8  &            &     1,810  &     1,810  &   1,810.0  &       8.6  \\

      c205 &     1,810  &     1,810  &   1,810.0  &       5.7  &            &     1,810  &     1,810  &   1,810.0  &       5.8  &            &     1,810  &     1,810  &   1,810.0  &       5.4  \\

      c206 &     1,810  &     1,810  &   1,810.0  &       6.0  &            &     1,810  &     1,810  &   1,810.0  &       6.3  &            &     1,810  &     1,810  &   1,810.0  &       5.7  \\

      c207 &     1,810  &     1,810  &   1,810.0  &       6.3  &            &     1,810  &     1,810  &   1,810.0  &       6.0  &            &     1,810  &     1,810  &   1,810.0  &       6.0  \\

      c208 &     1,810  &     1,810  &   1,810.0  &       6.6  &            &     1,810  &     1,810  &   1,810.0  &       6.2  &            &     1,810  &     1,810  &   1,810.0  &       6.0  \\
\hline
      r101 &     1,386  &     1,345  &   1,329.0  &      36.0  &            &     1,386  &     1,266  &   1,255.4  &      45.8  &            &     1,386  &     1,302  &   1,286.2  &      30.3  \\

      r102 &     1,420  &     1,412  &   1,407.8  &      31.5  &            &     1,420  &     1,371  &   1,367.5  &      43.9  &            &     1,420  &     1,387  &   1,383.0  &      37.1  \\

      r103 &     1,458  &     1,458  &   1,455.7  &      31.5  &            &     1,458  &     1,444  &   1,444.0  &      25.3  &            &     1,458  &     1,451  &   1,449.6  &      70.8  \\

      r104 &     1,458  &     1,458  &   1,458.0  &       5.3  &            &     1,458  &     1,458  &   1,458.0  &       5.9  &            &     1,458  &     1,458  &   1,458.0  &       7.2  \\

      r105 &     1,458  &     1,450  &   1,441.3  &      33.1  &            &     1,458  &     1,438  &   1,429.0  &      37.3  &            &     1,458  &     1,419  &   1,410.8  &      38.9  \\

      r106 &     1,458  &     1,458  &   1,458.0  &       5.4  &            &     1,458  &     1,458  &   1,458.0  &      22.1  &            &     1,458  &     1,458  &   1,456.2  &      14.3  \\

      r107 &     1,458  &     1,458  &   1,458.0  &       5.2  &            &     1,458  &     1,458  &   1,458.0  &       5.9  &            &     1,458  &     1,458  &   1,458.0  &       6.3  \\

      r108 &     1,458  &     1,458  &   1,458.0  &       6.1  &            &     1,458  &     1,458  &   1,458.0  &       6.8  &            &     1,458  &     1,458  &   1,458.0  &       7.3  \\

      r109 &     1,458  &     1,458  &   1,458.0  &       4.2  &            &     1,458  &     1,458  &   1,458.0  &       6.3  &            &     1,458  &     1,458  &   1,458.0  &       9.4  \\

      r110 &     1,458  &     1,458  &   1,458.0  &       4.5  &            &     1,458  &     1,458  &   1,458.0  &       5.7  &            &     1,458  &     1,458  &   1,458.0  &       5.9  \\

      r111 &     1,458  &     1,458  &   1,458.0  &       5.3  &            &     1,458  &     1,458  &   1,458.0  &       6.1  &            &     1,458  &     1,458  &   1,458.0  &       6.2  \\

      r112 &     1,458  &     1,458  &   1,458.0  &       5.9  &            &     1,458  &     1,458  &   1,458.0  &       6.7  &            &     1,458  &     1,458  &   1,458.0  &       8.0  \\
\hline
      r201 &     1,458  &     1,458  &   1,458.0  &       6.3  &            &     1,458  &     1,458  &   1,458.0  &       6.7  &            &     1,458  &     1,458  &   1,458.0  &       6.8  \\

      r202 &     1,458  &     1,458  &   1,458.0  &       7.1  &            &     1,458  &     1,458  &   1,458.0  &       7.1  &            &     1,458  &     1,458  &   1,458.0  &       7.4  \\

      r203 &     1,458  &     1,458  &   1,458.0  &       7.6  &            &     1,458  &     1,458  &   1,458.0  &       8.0  &            &     1,458  &     1,458  &   1,458.0  &       8.4  \\

      r204 &     1,458  &     1,458  &   1,458.0  &       9.2  &            &     1,458  &     1,458  &   1,458.0  &       9.3  &            &     1,458  &     1,458  &   1,458.0  &      10.3  \\

      r205 &     1,458  &     1,458  &   1,458.0  &       7.7  &            &     1,458  &     1,458  &   1,458.0  &       8.4  &            &     1,458  &     1,458  &   1,458.0  &       8.4  \\

      r206 &     1,458  &     1,458  &   1,458.0  &       8.0  &            &     1,458  &     1,458  &   1,458.0  &       9.1  &            &     1,458  &     1,458  &   1,458.0  &       8.8  \\

      r207 &     1,458  &     1,458  &   1,458.0  &       9.7  &            &     1,458  &     1,458  &   1,458.0  &       9.9  &            &     1,458  &     1,458  &   1,458.0  &      10.1  \\

      r208 &     1,458  &     1,458  &   1,458.0  &      11.3  &            &     1,458  &     1,458  &   1,458.0  &      11.2  &            &     1,458  &     1,458  &   1,458.0  &      12.3  \\

      r209 &     1,458  &     1,458  &   1,458.0  &       8.6  &            &     1,458  &     1,458  &   1,458.0  &       8.5  &            &     1,458  &     1,458  &   1,458.0  &       9.1  \\

      r210 &     1,458  &     1,458  &   1,458.0  &       7.9  &            &     1,458  &     1,458  &   1,458.0  &       8.4  &            &     1,458  &     1,458  &   1,458.0  &       8.6  \\

      r211 &     1,458  &     1,458  &   1,458.0  &      11.3  &            &     1,458  &     1,458  &   1,458.0  &      11.5  &            &     1,458  &     1,458  &   1,458.0  &      11.9  \\
\hline
     rc101 &     1,724  &     1,703  &   1,691.1  &      29.2  &            &     1,724  &     1,677  &   1,654.5  &      40.8  &            &     1,724  &     1,649  &   1,637.7  &      39.1  \\

     rc102 &     1,724  &     1,724  &   1,720.0  &      27.0  &            &     1,724  &     1,724  &   1,712.3  &      26.3  &            &     1,724  &     1,721  &   1,709.9  &      24.3  \\

     rc103 &     1,724  &     1,724  &   1,724.0  &       4.4  &            &     1,724  &     1,724  &   1,724.0  &       5.5  &            &     1,724  &     1,724  &   1,724.0  &       6.0  \\

     rc104 &     1,724  &     1,724  &   1,724.0  &       5.0  &            &     1,724  &     1,724  &   1,724.0  &       6.0  &            &     1,724  &     1,724  &   1,724.0  &       6.4  \\

     rc105 &     1,724  &     1,724  &   1,717.8  &      22.8  &            &     1,724  &     1,709  &   1,702.9  &      28.2  &            &     1,724  &     1,701  &   1,692.8  &      36.0  \\

     rc106 &     1,724  &     1,724  &   1,724.0  &       5.9  &            &     1,724  &     1,724  &   1,724.0  &       7.5  &            &     1,724  &     1,724  &   1,724.0  &      25.6  \\

     rc107 &     1,724  &     1,724  &   1,724.0  &       4.4  &            &     1,724  &     1,724  &   1,724.0  &       6.2  &            &     1,724  &     1,724  &   1,724.0  &       5.7  \\

     rc108 &     1,724  &     1,724  &   1,724.0  &       4.9  &            &     1,724  &     1,724  &   1,724.0  &       5.9  &            &     1,724  &     1,724  &   1,724.0  &       6.1  \\
\hline
     rc201 &     1,724  &     1,724  &   1,724.0  &       5.8  &            &     1,724  &     1,724  &   1,724.0  &       6.2  &            &     1,724  &     1,724  &   1,724.0  &       5.9  \\

     rc202 &     1,724  &     1,724  &   1,724.0  &       6.4  &            &     1,724  &     1,724  &   1,724.0  &       6.5  &            &     1,724  &     1,724  &   1,724.0  &       5.4  \\

     rc203 &     1,724  &     1,724  &   1,724.0  &       6.4  &            &     1,724  &     1,724  &   1,724.0  &       6.6  &            &     1,724  &     1,724  &   1,724.0  &       6.5  \\

     rc204 &     1,724  &     1,724  &   1,724.0  &       8.2  &            &     1,724  &     1,724  &   1,724.0  &       7.8  &            &     1,724  &     1,724  &   1,724.0  &       9.1  \\

     rc205 &     1,724  &     1,724  &   1,724.0  &       5.9  &            &     1,724  &     1,724  &   1,724.0  &       5.8  &            &     1,724  &     1,724  &   1,724.0  &       4.9  \\

     rc206 &     1,724  &     1,724  &   1,724.0  &       6.5  &            &     1,724  &     1,724  &   1,724.0  &       6.9  &            &     1,724  &     1,724  &   1,724.0  &       5.7  \\

     rc207 &     1,724  &     1,724  &   1,724.0  &       7.3  &            &     1,724  &     1,724  &   1,724.0  &       6.3  &            &     1,724  &     1,724  &   1,724.0  &       6.2  \\

     rc208 &     1,724  &     1,724  &   1,724.0  &       8.9  &            &     1,724  &     1,724  &   1,724.0  &       8.5  &            &     1,724  &     1,724  &   1,724.0  &       7.3  \\
\hline
\end{tabular}
}
\end{center}
\end{table*}

Theoretically speaking, the optimal solution value of some instance with $w = d$ must be greater than or equal to that of the same instance with $w = k d$, where $k$ is an integral number. This is because we can always construct a feasible solution to an instance with $w = d$ from any solution to this instance with $w = kd$. For example, the optimal solution value of an instance with $w = 1$ must be greater than those of this instance with $w = 3$ and $w =5$. Since our tabu search algorithm is a stochastic approach, it is possible that the maximum workload of some instance with $w = 3$ or $w =5$ is larger than that of this instance with $w = 1$. Fortunately, we did not encounter such phenomenon in our experiments. However, when $w_2/w_1$ is not an integer, an instance with $w_2$ may have larger optimal solution value than this instance with $w_1$. For example, the optimal solution value of an instance with $w=5$ may be larger than this instance with $w = 3$. In these tables, we can find several instances with $w =3$ have larger maximum workload than their counterparts with $w = 5$. Since these maximum workloads may not be optimal, we cannot judge whether these phenomena were resulted from the randomness of the tabu search algorithm or the nature of the instances. Obviously, for each instance with a certain $w$, the maximum workload increases as the number of vehicles.

For each instance group, we calculated the average of all ``Ave. Time'' and show the statistical results in Table \ref{tab:t1}.  From this table, we can observe that in most cases (those are not marked in bold), the average computation times increase as the value of $w$.
\renewcommand{\arraystretch}{1.0}
\begin{table}[!h]
  \centering
  \caption{The average computation time of each MPISP instance group.}
\subtable[]{
\begin{small}
    \begin{tabular}{cccc}
    \hline
    $m = 7$   & $w = 1$   & $w = 3$   & $w = 5$ \\
    \hline
    c1    & 9.31  & 12.58  & 23.07  \\
    c2    & 15.74  & 16.59  & 20.29  \\
    r1    & 51.16  & 74.75  & 88.86  \\
    r2    & 8.76  & 10.07  & 11.17  \\
    rc1   & 51.14  & 70.69  & 104.95  \\
    rc2   & 13.84  & 17.20  & 18.23  \\
    \hline
    \end{tabular}%
    \end{small}
    \label{tab:s1}
}
\qquad
\subtable[]{
\begin{small}
    \begin{tabular}{cccc}
    \hline
    $m = 9$   & $w = 1$   & $w = 3$   & $w = 5$ \\
    \hline
    {\bf c1}    & 26.22  & 26.23  & 11.32  \\
    c2    & 5.26  & 5.34  & 5.36  \\
    r1    & 55.88  & 77.30  & 91.78  \\
    r2    & 7.95  & 8.12  & 8.67  \\
    rc1   & 47.48  & 72.88  & 94.29  \\
    {\bf rc2}   & 6.23  & 5.94  & 5.54  \\
    \hline
    \end{tabular}%
    \end{small}
}
\qquad
\subtable[]{
\begin{small}
    \begin{tabular}{cccc}
    \hline
    $m = 11$  & $w = 1$   & $w = 3$   & $w = 5$ \\
    \hline
    c1    & 5.17  & 7.00  & 8.37  \\
    c2    & 5.91  & 5.98  & 6.06  \\
    r1    & 20.16  & 33.95  & 49.82  \\
    r2    & 8.24  & 8.47  & 8.98  \\
    rc1   & 28.21  & 36.71  & 48.16  \\
    {\bf rc2}   & 6.64  & 6.35  & 6.04  \\
    \hline
    \end{tabular}%
    \end{small}
}
\qquad
\subtable[]{
\begin{small}
    \begin{tabular}{cccc}
    \hline
    $m = 13$  & $w = 1$   & $w = 3$   & $w = 5$ \\
    \hline
    {\bf c1}    & 5.50  & 5.10  & 5.30  \\
    {\bf c2}    & 6.34  & 6.20  & 6.18  \\
    r1    & 14.50  & 18.15  & 20.14  \\
    r2    & 8.61  & 8.92  & 9.28  \\
    rc1   & 12.95  & 15.80  & 18.65  \\
    {\bf rc2}   & 6.93  & 6.83  & 6.38  \\
    \hline
    \end{tabular}%
    \end{small}
}
  \label{tab:t1}%
\end{table}%

\section{Conclusion}
\label{sec:con}
This paper introduces an inspector scheduling problem in which a set of inspectors are dispatched to complete a set of inspection requests at different locations in a multi-period planning horizon. At the end of each period, each inspector is not required to return to the depot but has to stay at one of the inspection locations for recuperation. We first studied the way of computing the shortest transit time between any pair of locations when the working time periods are considered. Next, we introduced several local search operators that were adapted from classical VRPTW operators and integrated these adapted operators in a tabu search framework. Moreover, we presented a constrained knapsack model that is able to produce an upper bound for the MPISP. Finally, we evaluated the algorithm based on 370 TOPTW instances and 672 MPISP instances. The experimental results reported in this study show the effectiveness of our algorithm and can serve as benchmarks for future researchers. Since the working time windows of the scheduling subjects and the use of waypoint are very common considerations in practice, a possible research direction can focus on studying the variants of other existing vehicle routing models that involves these two factors.

\section*{Acknowledgments}
This research was partially supported by the Fundamental Research Funds for the Central Universities, HUST (Grant No. 2014QN206) and National Natural Science Foundation of China (Grant No. 71201065).








\end{document}